\documentclass[journal]{IEEEtran}
\usepackage[utf8]{inputenc} 
\usepackage[T1]{fontenc}    
\usepackage{lineno, hyperref}       
\usepackage{url}            
\usepackage{booktabs}       
\usepackage{amsfonts}       
\usepackage{nicefrac}       
\usepackage{microtype}      
\usepackage{amsmath, amssymb, amsthm, amsfonts}
\usepackage{multirow}
\usepackage{threeparttable}
\usepackage{graphicx}
\usepackage{caption}
\usepackage{subfigure}
\usepackage{wrapfig}
\usepackage{color}
\usepackage{diagbox}
\usepackage{latexsym}
\usepackage{makecell}
\usepackage{threeparttable}
\usepackage{comment}

\ifCLASSINFOpdf
\else
\fi
\begin{document}
%
\title{Graph-based Visual-Semantic Entanglement Network for Zero-shot Image Recognition}
%
%
%

\author{Yang~Hu,~\IEEEmembership{Student Member,~IEEE,}
        Guihua~Wen,
        Adriane~Chapman,
        Pei~Yang,~\IEEEmembership{Member,~IEEE,}
        Mingnan~Luo,
        Yingxue~Xu,
        Dan~Dai,
        Wendy~Hall 
\thanks{Yang~Hu is with South China University of Technology, China, University of Southampton, UK, e-mail: superhy199148@hotmail.com, Guihua~Wen, Mingnan~Luo, Pei~Yang, and Yingxue~Xu are with South China University of Technology, China, Adriane~Chapman and Wendy~Hall are with University of Southampton, UK, and Dan~Dai is with South China University of Technology, China, University of Lincoln, UK.}
\thanks{Guihua~Wen and Wendy Hall are the Corresponding Authors, e-mail: crghwen@scut.edu.cn, wh@ecs.soton.ac.uk.}
}

%
%

\markboth{Journal of \LaTeX\ Class Files,~Vol.~X, No.~X, August~2020}%
{Shell \MakeLowercase{\textit{et al.}}: Bare Demo of IEEEtran.cls for IEEE Journals}
%



\maketitle

\begin{abstract}
Zero-shot learning uses semantic attributes to connect the search space of unseen objects. In recent years, although the deep convolutional network brings powerful visual modeling capabilities to the ZSL task, its visual features have severe pattern inertia and lack of representation of semantic relationships, which leads to severe bias and ambiguity. In response to this, we propose the Graph-based Visual-Semantic Entanglement Network to conduct graph modeling of visual features, which is mapped to semantic attributes by using a knowledge graph, it contains several novel designs: 1. it establishes a multi-path entangled network with the convolutional neural network (CNN) and the graph convolutional network (GCN), which input the visual features from CNN to GCN to model the implicit semantic relations, then GCN feedback the graph modeled information to CNN features; 2. it uses attribute word vectors as the target for the graph semantic modeling of GCN, which forms a self-consistent regression for graph modeling and supervise GCN to learn more personalized attribute relations; 3. it fuses and supplements the hierarchical visual-semantic features refined by graph modeling into visual embedding. Our method outperforms state-of-the-art approaches on multiple representative ZSL datasets: AwA2, CUB, and SUN by promoting the semantic linkage modelling of visual features.
\end{abstract}

\begin{IEEEkeywords}
Zero-shot learning, visual-semantic modeling, graph convolutional network, semantic knowledge graph, attribute word embedding.
\end{IEEEkeywords}

%
\IEEEpeerreviewmaketitle

\section{Introduction}
\label{sec1}

\IEEEPARstart{Z}{ero}-shot learning (ZSL) presents a new paradigm for image recognition, whose goal is to train models to recognize categories that have never been seen before~\cite{lampert2009learning}. Because many instances in datasets of the real world do not have associated classes, ZSL is required to make computer vision (CV) more generalized~\cite{kodirov2017semantic,xie2019attentive,kampffmeyer2019rethinking,jiang2019transferable,ni2019dual}. In typical supervised learning, some samples are unlabelled. However, in ZSL, the unlabelled are some entire categories (unseen), and the only clue that connects the seen and unseen classes is the structured attribute assigned to each class. In recent years, architectures using deep convolutional networks (CNN) are used as the ZSL embedding subnet~\cite{song2018transductive,ye2019progressive,li2018discriminative,jiang2019transferable}. CNN based ZSL models are usually pre-trained by ImageNet~\cite{russakovsky2015imagenet} to earn the initial visual perception ability~\cite{song2018transductive,ye2019progressive,zhu2019semantic}. However, such models only provide mappings between classes and observed patterns; knowledge about relationships within the attributes of seen and unseen classes are not utilized.

\begin{figure}[t]
\centering
\includegraphics[scale=0.3]{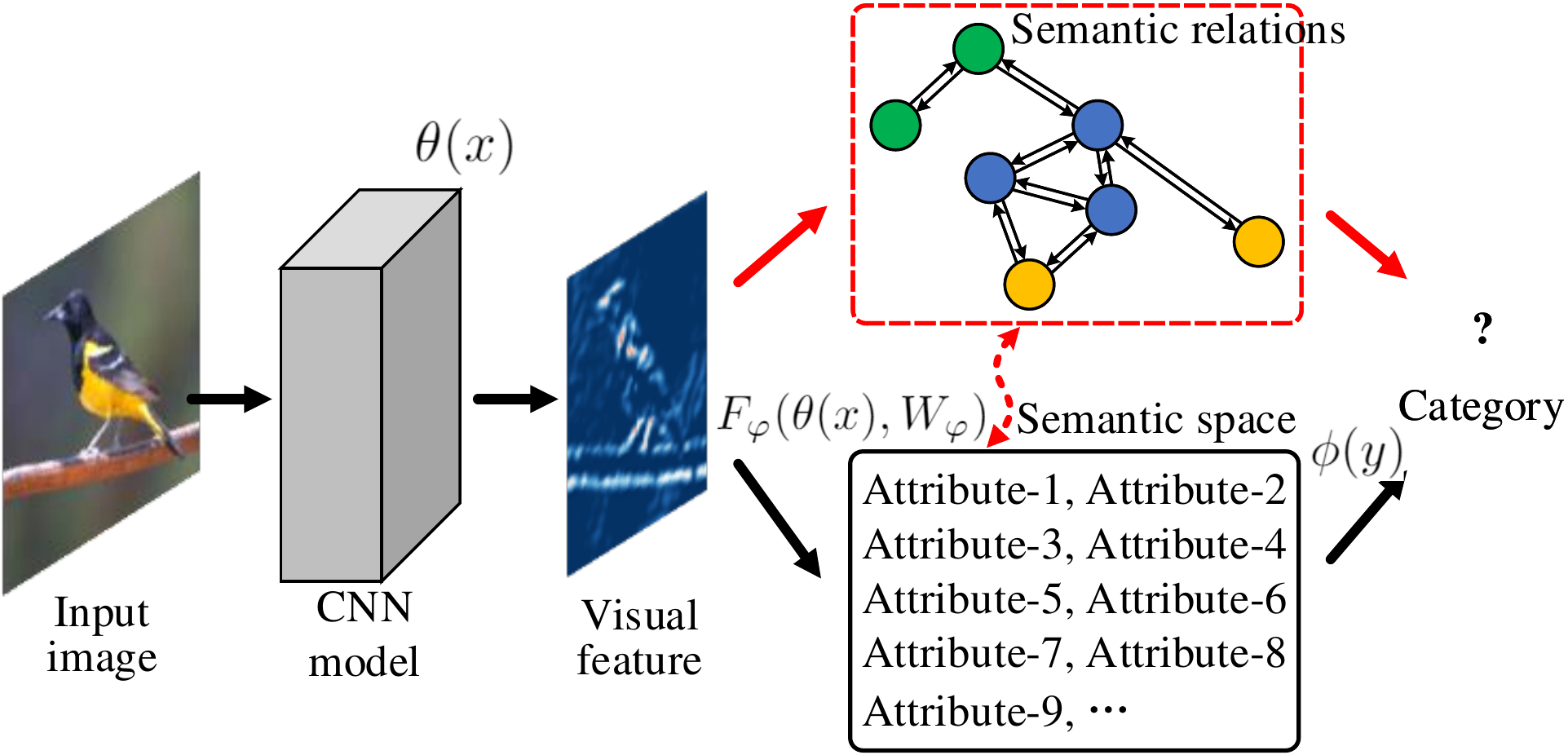}
\caption{Illustrating the strategy of graph-based visual-semantic embedding. The red lines and dashed box represent the modelling of implicit semantic links in visual features and help visual-semantic bridging complete the category mapping.}
\label{fig1-1}
\end{figure}

The components outlined in black in Figure~\ref{fig1-1} shows how current ZSL model uses attributes. In existing ZSL methods, there are two types of settings: the inductive setting methods that have no access to unseen classes~\cite{zhang2015zero,romera2015embarrassingly,zhu2019semantic} and transductive setting methods~\cite{kumar2018generalized,song2018transductive,wan2019transductive} that have access to unseen samples without labels. However, in both the inductive and transductive settings, visual image modeling and attribute mapping modeling are separate. For instance, in a ZSL task, the input image is denoted as $x$. The visual embedding function formed by CNN is $\theta (\cdot)$. Separated from this, the semantic embeddings $\phi (y)$ indicate the distribution vector of the attributes to a class $y$. $\mathcal{F} (x, y, W_{\varphi}) = \mathcal{F}_{\varphi} (\theta (x), W_{\varphi}) \phi (y)$ is the score function of the classification. The only connection between the visual and the semantic embeddings is the visual-semantic bridging $\mathcal{F}_{\varphi} (\cdot, W_{\varphi})$, which is usually constructed by a few fully-connected layers (FC). Facing visual embedding $\theta (\cdot)$ with strong pattern inertia, $\mathcal{F}_{\varphi} (\cdot, W_{\varphi})$ bears an excessive modelling pressure and hard to reverse prediction bias. Although the latent attributes mechanism reduces the skew caused by the attribute inertia~\cite{li2018discriminative,liu2019attribute}, the existing ZSL models still ignore the implicit semantic linkages in visual features.

\begin{figure}[t]
\centering
\includegraphics[scale=0.6]{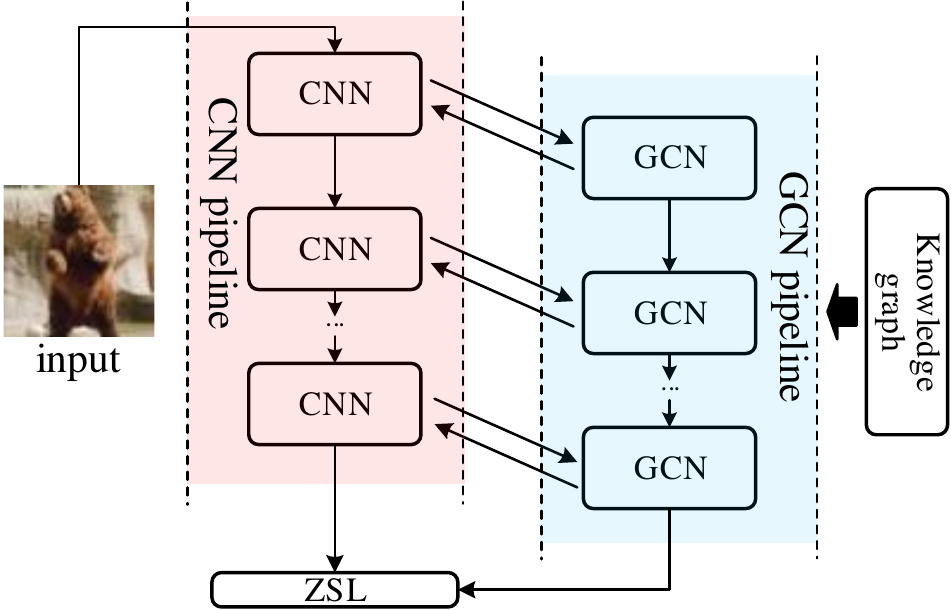}
\caption{Illustrating the structure of Graph-based Visual-Semantic Entanglement. The left is the CNN pipeline, which is used to model the visual feature of the input image, and the right is the GCN pipeline, which is used to model the semantic relationship in the visual features and provide feedback. A pre-set knowledge graph supports the GCN pipeline.}
\label{fig1-2}
\end{figure}

In order to address the problem of typical visual features which lack knowledge about relationships within the attributes, we must utilize knowledge of the attribute connections to complement the mapping, modeling, and fusion of implicit semantic relationships, as shown in red in Figure~\ref{fig1-1}. To do this, we propose the \textbf{Graph-based Visual-Semantic Entanglement} (GVSE) network. As illustrated in Figure~\ref{fig1-2}, GVSE network mainly proposes the visual-semantic entanglement structure: The concept in the attribute space is closer to the semantics in the image representation. Attribute-based graph modeling can directly interact with semantic information in visual features. To utilize the relationships amongst attributes found in the semantic knowledge graph, we build the architecture with CNN and graph convolutional network (GCN)~\cite{kipf2016semi} two entangled pipelines, we map CNN visual features to attribute-based knowledge graphs, then utilize GCN to model the implicit semantic relations of intermediate visual features and feedback to the CNN. GVSE network also contains some effective designs: 1. To provide better sets of characterized attribute relations for the model, we create a self-consistent system with GCN modelling by creating attribute word vectors to achieve semantic regression. 2. To integrate this new semantic-derived information into the classic ZSL model, we skip-connect the graph modeled visual features to the visual embedding $\theta (x)$. The purpose is to fuse the graph enhanced features to visual embedding and help the loss back-propagation from semantic embedding $\phi (y)$.

The contributions of our study can be summarised as:

1. We propose the \textbf{Graph-based Visual-Semantic Entanglement} network, in which, GCN remodels the CNN visual features with an pre-set knowledge graph contains attribute relational information, and feedbacks semantic relations to the deep visual CNN pipeline.

2. We design to utilize the attribute word vectors as the regression output of the GCN pipeline, which offers a self consistent supervision for semantic graph modelling and helps the model to mine more personalized attribute relationships based on the initialized attribute knowledge graph.

3. We propose to build the skip-connection from GCN block to visual embedding, which combines the graph remodeled features into visual embedding to enhance the semantic relational representation before we use it in ZSL prediction.

4. We evaluate the proposed GVSE network and supporting techniques. The proposed method achieves state-of-the-art performance on ZSL datasets: AwA2~\cite{xian2018zero}, CUB~\cite{WahCUB_200_2011}, and SUN~\cite{patterson2012sun}. We perform ablation studies and visualization analysis to validate the effectiveness of our designs.

\section{Related Works}
\label{sec2}

\subsection{Zero-shot Learning}
\label{sec2.1}

Zero-shot learning (ZSL) uses semantic attributes to connect the search space of unseen objects. The initiatory ZSL uses a two-stage method; the first stage predicts the attribute while the second stage maps the attribute to the class, based on direct attribute prediction (DAP)~\cite{lampert2009learning}. The prediction of ZSL relies on a shared semantic space which can be user-defined attribute annotations~\cite{farhadi2009describing} or unsupervised word embeddings~\cite{mikolov2013distributed}.

To achieve targets: 1. removing the gap between attribute predictions to the final category; 2. correcting assumptions that attributes are independent of each other, ESZSL~\cite{romera2015embarrassingly} and SCoRe~\cite{morgado2017semantically} propose to use a multi-layer neural network to train the visual-semantic bridge. SAE~\cite{kodirov2017semantic} introduces external word embeddings as an assistant. However, the "\textit{domain shift}"~\cite{fu2015transductive} shows that different categories have distinct visual expressions on some attributes. It causes severe category bias, especially in generalized ZSL setting~\cite{chao2016empirical}, which mixes the seen and unseen classes in a  search space. Methods~\cite{jiang2019transferable,xu2019zero} add regulation in the loss function. AEZSL~\cite{niu2019zero} generates embedding spaces from multiple perspectives, and PREN~\cite{ye2019progressive} uses the ensemble method to fuse them. The generative models like adversarial learning~\cite{ni2019dual,xie2019attentive,yu2020episode} and variational auto-encoder~\cite{xian2019f,gao2020zero} are also applied in ZSL to generate more vivid embedding spaces. However, such models usually have inadequate training stability, which has a high parameter adjustment requirement for researchers and is not convenient for application to more fields. The attention-based methods highlight detailed parts. However, they still miss the semantic relations~\cite{zhu2019semantic,huynh2020fine}. The transductive ZSL methods~\cite{song2018transductive,ye2019progressive,wan2019transductive,zhang2019hierarchical} even learn a part of unlabeled unseen class samples to remedy the natural lacking of inference clues in visual-semantic modelling. Moreover, LDF~\cite{li2018discriminative} suggests that human-defined attributes cannot fully depict semantic information and propose the latent attribute mechanism to learn the pseudo-attributes automatically. LFGAA~\cite{liu2019attribute} adds attention values for both human-defined and pseudo attributes. AMS-SFE~\cite{guo2020novel} uses manifold modeling on the expanded semantic feature.

Instead, we argue that another vital reason for the "domain shift" is that the connection between visual features and semantic space is too weak. Specifically, visual features and semantic space are independent of each other, and the semantic connection between visual features is not reflected. The GVSE has obvious differences with the existing ZSL methods: 1. The embedding subnet in previous works focuses on modelling the visual features and ignores the implicit semantic linkages in deep visual features; 2. The GVSE network introduces the unbiased knowledge graph to supervise the visual modelling and uses the GCN to learn the cognition of semantic relations.

Some methods also use external semantic ontology~\cite{kodirov2017semantic,morgado2017semantically} or knowledge graph~\cite{wang2018zero,kampffmeyer2019rethinking,chenzero2020zero,liu2020attribute} to assist ZSL models. However, they are still different from GVSE: 1. The graph in all these works is based on the category relations, while we use attributes as nodes to model the attribute relationship. We consider that attribute-based graph modeling can directly map with the latent semantics in visual features. Besides, SEKG~\cite{wang2018zero} and DGP~\cite{kampffmeyer2019rethinking} rely on the topology of pre-built ontology WordNet; 2. They use the tandem structure to combine GCN graph modeling and CNN visual modeling modules, which cannot feed the semantic graph information back to the visual features. From another perspective, the intermediate visual features have a rare chance to be modeled on the semantic knowledge graph.

\subsection{Graph Neural Network for Computer Vision}
\label{sec2.2}

Graph neural networks (GNNs) are widely applied in computer vision due to their excellent modelling capabilities for non-Euclidean relations~\cite{bronstein2017geometric,kipf2016semi}. The benefit of visual modelling applications includes: various recognition tasks~\cite{chen2018knowledge,chen2020knowledge},  activity recognition~\cite{lu2019gaim}, object detection~\cite{sun2019relational} and segmentation~\cite{zhang2019dual}, as well as visual point cloud data modelling~\cite{li2019deepgcns}. Researchers use a GNN to model the links between category labels~\cite{chen2019learning,chenzero2020zero,gao2020ci} or integrate external knowledge~\cite{chen2018knowledge,cui2017general}. Graph learning also enhances the representation of the connections between visual objects and semantic entities and improves detection and segmentation~\cite{zhang2019dual}. Visual-semantic interaction tasks like visual question \& answer (VQA) and image caption are emphasised application areas of a GNN~\cite{li2019visual,wang2020learning,li2019know},  though which the logic relations between visual features are modeled, which gives the model semantic reasoning ability and promotes the info-transmission of visual signals to semantic.

A graph convolution~\cite{kipf2016semi} is the most common method of graph learning. Also, abundant studies can confirm that implicit semantic linkages exist in visual features~\cite{liu2020multi,li2019know}. Graph modeling is a powerful means of loading structured knowledge into deep models, especially for the ZSL model with a great demand for semantic relationship modeling.

The above methods are significantly different from the proposed GVSE network: 1. In the existing studies, GNN models the outcomes of a CNN or external structured knowledge without modelling intermediate visual features; 2. The previous GNN-based models lack the ability to feed the graph modelled features back the visual representation; 3. Visual modeling and graph modeling in the existing studies are concatenated in series; the previous part modelling bias will easily affect subsequent modeling. By contrast, the GVSE method uses a GCN to reconstruct and readjust deep visual representations, which uses parallel pipelines to model visual and semantic features separately, with sufficient feedback and robust interactivity. It enhances the ability to model semantic relationships in visual models while retaining the original visual representation.

\section{Proposed Method}
\label{sec3}

\subsection{Problem Formulation and Notations}
\label{sec3.1}

The ZSL task is formulated as follows: there is a seen dataset $\mathcal{S} = \{ (x_{i}^{s}, y_{i}^{s}) \}_{i=1}^{N^{s}}$ which contains $N^{s}$ samples for training, where $x_{i}^{s}$ denotes the $i$-th image and $y_{i}^{s} \in \mathcal{Y}^{S}$ is the category label of it. There is an another unseen dataset $\mathcal{U} = \{ (x_{i}^{u}, y_{i}^{u}) \}_{i=1}^{N^{u}}$ with similar form. The seen and unseen category sets $\mathcal{Y}^{\mathcal{S}}$ and $\mathcal{Y}^{\mathcal{U}}$ obey the following constraints: $\mathcal{Y}^{S} \cap \mathcal{Y}^{U} = \emptyset, \mathcal{Y}^{S} \cup \mathcal{Y}^{U} = \mathcal{Y}$ where $\mathcal{Y}$ is the total category set. $\mathcal{Y}^{S}$ and $\mathcal{Y}^{U}$ share a semantic attribute space: $ \forall y_{i} \ \exists <Att_{1}, \dots, Att_{m}>$ as the only bridge between them and $y_{i} \in \mathcal{Y}$. $Att_i$ refers to the attribute which is usually a word or concept, and $m$ is the number of attributes. The goal of conventional ZSL (CZSL) is to learn the classifier with the search space of unseen classes $\mathcal{Y}^{\mathcal{U}}$, for more challenging generalized ZSL (GZSL), the search space of expected classifier is $\mathcal{Y}$.

Table~\ref{table3-r1} lists some notations and their interpretations that specifically appear in this paper.

\begin{table}[t]
\scriptsize
\begin{center}
\caption{Explanation of the special notations utilized in this paper}
\label{table3-r1}
\begin{tabular}{c|c}
\hline
Notation & Meaning \\
\hline
$\mathcal{N} (\cdot)$ & The normalization function within $[0,1]$ \\
\hline
$(\cdot)^{\top}$ & Matrix transpose \\
\hline
$<\cdot, \cdot>$ & A collaborative operating pair consisting of two elements \\
\hline
$[\sim, l]$ & For a sequence, all elements from $1$ to $l$ \\
\hline
$[l, \sim]$ & For a sequence, all elements from $l$ to the end \\
\hline
$[\sim]$ & All elements in a sequence \\
\hline
$\Join$ & The concatenation operation \\
\hline
$[\cdot]_{+}$ & Equivalent to $\mathrm{max}(0, \cdot)$ \\
\hline
\end{tabular}
\end{center}
\end{table}

\subsection{Overview Framework}
\label{sec3.2}

\begin{figure*}[t]
\centering
\includegraphics[scale=0.6]{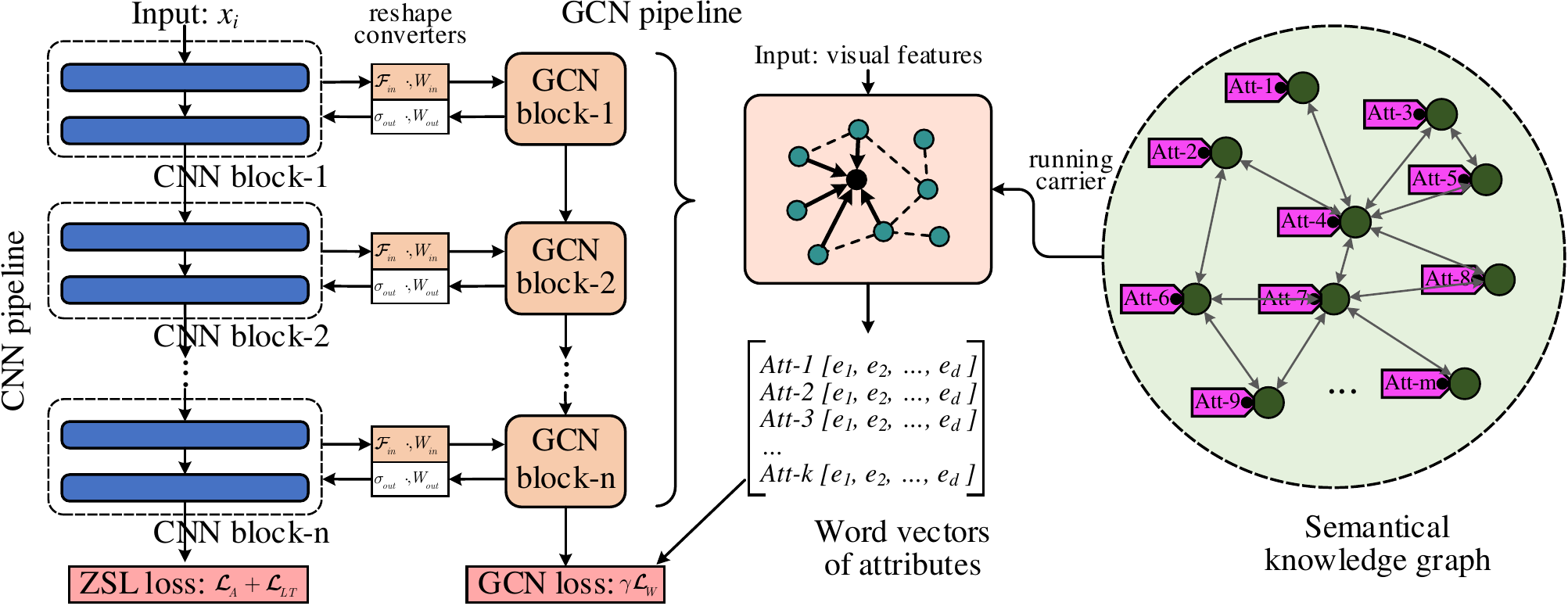}
\caption{Overview framework of the GVSE network. The leftmost area is the CNN pipeline, which performs visual modeling for the images. The middle shows the GCN pipeline, which models the implicit semantic relationships in visual features. The reshape converters regular the dimension of visual feature input and semantic information feedback, which are used to support dual-pipe interaction. The rightmost area is the semantic knowledge graph, the vertexes of which refer to the attributes. It is the "running carrier" for GCN pipeline: the graph convolution algorithm is running based on the semantic knowledge graph.}
\label{fig3-1}
\end{figure*}

In order to achieve the goal of visual-semantic entangled feature modeling for ZSL, the following steps must be taken:

1. \textbf{Construction of Semantic Knowledge Graph:} Construct the semantic knowledge graph for the execution of GCN. The knowledge graph extracts co-occurrence relations of attributes.

2. \textbf{Establish the Visual-Semantic Dual-pipe Network Structure:} Establish a dual-pipe network structure with clear responsibilities as Figure~\ref{fig1-2}, in which CNN is responsible for conventional image visual modeling, and GCN is responsible for graph modeling of semantic relations of visual features.

3. \textbf{Design the Entanglement Strategy of CNN and GCN:} Establish the interaction functions between CNN visual modeling and GCN semantic modeling. GCN receives visual features from CNN as input, and CNN gets GCN semantic information to further optimize visual features.

4. \textbf{Fusion of Graph Semantic Encoding and Visual Representation:} In order to further strengthen the previous feature representation of the ZSL bridging, we merge the semantic graph modeled features from the GCN blocks into the final visual embedding.

Figure~\ref{fig3-1} shows the overview framework of the GVSE network, which illustrates the semantic knowledge graph for attributes, CNN visual modeling pipeline, GCN semantic modeling pipeline, the target outputs of dual-pipe, and the interaction support modules for dual-pipe entanglement.

\subsubsection{Semantic Knowledge Graph}
\label{sec3.2.1}

\begin{figure}[t]
\centering
\includegraphics[scale=0.6]{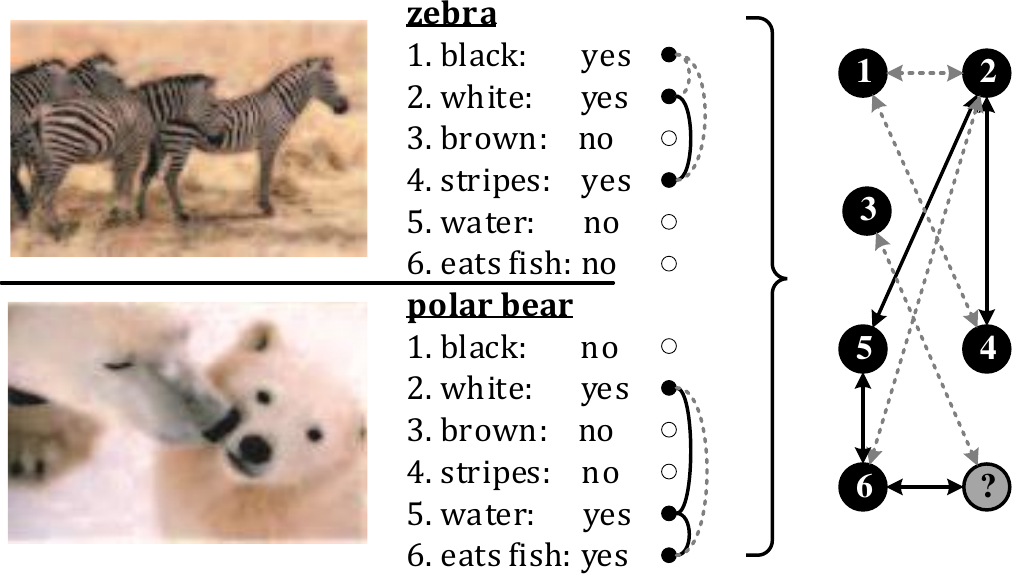}
\caption{Illustration of attribute linkage construction. The gray-dotted lines are the co-occurrence relations with low $\mathcal{PMI}$ value, and the black lines indicate a high $\mathcal{PMI}$ value between attributes. There is no separate sub-graph in our knowledge graph, and '?' represents another attribute vertex that connects to the appeared attributes.}
\label{fig3-2}
\end{figure}

Since GCN needs to run on a pre-defined graph structure, before constructing the dual-pipe visual-semantic neural network architecture, we first introduce the construction strategy of the semantic knowledge graph. This knowledge graph will have the following characteristics: 1. The visual features are directly mapped with attributes. Thus, the knowledge graph is expected can represent the fundamental relationship of attributes; 2. The establishment of the knowledge graph is off-line and convenient; 3. The establishment of the knowledge graph can apply to the various scenes, without being restricted by the ontology adapted to the data set.

Co-occurrence relationship is one of the basic linkages between attributes. It can be obtained by counting how often the attribute appears together in each category. Instead of applying WordNet as a prior knowledge graph~\cite{wang2018zero,kampffmeyer2019rethinking}, we construct the semantic knowledge graph based on attribute co-occurrence relations concerning all categories. The attributes with co-occurrence relations will have a high probability of existing in the same category's visual representation.

A knowledge graph, $\mathcal{G}_{att} = (\mathcal{V}, \mathcal{E})$, contains vertices $\mathcal{V} = \{ v_{1}, v_{2}, \dots, v_{m} \}$ and edges $\mathcal{E}$ between them. We utilize a symmetric matrix to encode edges $ \left[[l_{i,j}]\right]$ where $l_{i,j} = 1$ means there is a linkage between vertices $v_{i}$ and $v_{j}$, else not. The construction of our knowledge graph requires the attribute space indicates the unequivocal concepts. In our semantic knowledge graph, the attributes are used to define the vertices, and the point-wise mutual information (PMI)~\cite{bouma2009normalized} is used to calculate the attributes co-occurrence and determine the connection between attribute vertices, as follows:
\begin{equation}
\label{eq3-1}
\mathcal{PMI} \left( v_{i}, v_{j} \right) = \mathcal{N} \left( \mathrm{log} \frac{p \left( v_{i}, v_{j} \right)}{p \left( v_{i} \right)  p \left( v_{j} \right)} \right),
\end{equation}
where $\mathcal{PMI} \left( v_{i}, v_{j} \right)$ is the PMI between attributes $v_{i}$ and $v_{j}$, $p(v)$ is the occurrence probability of attribute $v$, and the co-occurrence probability $p (v_{i}, v_{j})$ is the ratio of the category amount with both $v_{i}$ and $v_{j}$ to the total number of categories. $\mathcal{N}$ denotes the normalization function within $[0, 1]$. Figure~\ref{fig3-1} illustrates the construction strategy of the semantic knowledge graph, and the edge is built between the vertices with a PMI of higher than the threshold $\delta$, as:
\begin{equation}
\label{eq3-2}
l_{i,j} = \left\{
\begin{array}{ll}
1, \quad \mathcal{PMI} \left( v_{i}, v_{j} \right) > \delta \\
0, \quad \mathrm{else}
\end{array}
\right.,
\end{equation}

Figure~\ref{fig3-2} shows an example of edge construction in our semantic knowledge graph. After obtaining the pre-designed semantic knowledge graph $\mathcal{G}_{att} = (\mathcal{V}, \mathcal{E})$, the dual-pipe network structure is introduced in the followed section.

\subsubsection{Dual-pipe Network Structure}
\label{sec3.2.2}

\begin{figure}[t]
\centering
\includegraphics[scale=0.65]{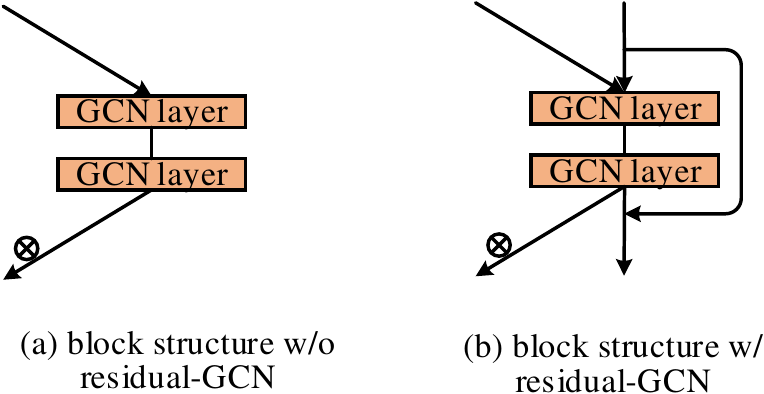}
\caption{Comparison of different GCN blocks. (a) The GCN blocks without residual structure; (b) The GCN blocks with residual skip-connection.}
\label{fig3-3}
\end{figure}

The two modeling pipelines we designed to have a clear division of labour. We first define the CNN pipeline for visual feature modeling as:
\begin{equation}
\label{eq3-3}
\theta(x_{i}) = \mathcal{F}_{conv} \left(x_{i}, W_{\theta} \right),
\end{equation}
where $x_{i}$ refers to the input and $\theta (x_{i})$ is the visual embedding modeled and provided by CNN pipeline $\mathcal{F}_{conv}$, which is parameterized by weights $W_{\theta}$. In our study, the CNN pipeline $\mathcal{F}_{conv}$ can be constructed with any of existing popular CNN backbone~\cite{simonyan2014very,he2016deep}.

The target prediction output of the CNN pipeline is the ZSL classification score $p(y | x_{i})$, and it is calculated as follows:
\begin{equation}
\label{eq3-4}
p(y | x_{i}) = \mathcal{F}_{\varphi} \left(\theta(x_{i}), W_{\varphi} \right)^{\top} \phi(y),
\end{equation}
where $\mathcal{F}_{\varphi}: \mathbb{R}^{d_v} \to \mathbb{R}^{m}$ is the attribute prediction layer with parameter $W_{\varphi} \in \mathbb{R}^{d_v \times m}$, and $\phi(y) \in \mathbb{R}^{m \times |\mathcal{Y}| }$ denotes the attribute distribution. $d_v$ is the dimension of visual features, $m$ and $|\mathcal{Y}|$ refer to the numbers of attributes and categories.

Based on the premise that there are implicit semantic linkages in visual features, and this has been found in other CV studies~\cite{wang2018non}, we design the GCN pipeline to model the semantic relations in visual features. As illustrated by Figure 3, note that we have feature maps $X^{(l)} = \mathcal{F}_{conv}^{[\sim, l]} \left(x_{i}, W_{\theta}^{[\sim, l]} \right)$ from the $l$-th block of CNN pipeline, where $[\sim, l]$ means "from $1$ to $l$". We use $X^{(l)}$ as the input of the corresponding GCN block $\mathcal{F}_{\mathcal{G}}^{(l)}$, as follows:
\begin{equation}
\label{eq3-5}
f_{\mathcal{G}}^{(l)} = \mathcal{F}_{\mathcal{G}}^{(l)} \left(X^{(l)}, <\mathcal{G}_{att}, W_{\mathcal{G}}^{(l)}> \right),
\end{equation}
where $f_{\mathcal{G}}^{(l)}$ is the output of the $l$-th GCN block, $W_{\mathcal{G}}^{(l)}$ is the parameters of $\mathcal{F}_{\mathcal{G}}^{(l)}$ that running on graph $\mathcal{G}_{att}$. The elements in $<\cdot, \cdot>$ must be utilized together. Each GCN block contains $2$ GCN layers, which is defined in~\cite{kipf2016semi}, as follows:
\begin{equation}
\label{eq3-6}
\mathcal{H}^{(i+1)} = \sigma \left(D^{-1} G_{a} \mathcal{H}^{(i)} {W}_{\mathcal{G}}^{(i)} \right)
\end{equation}
where $\mathcal{H}^{(i)}$ and $\mathcal{H}^{(i+1)}$ denotes any two layers in GCN, $D$ and $G_{a}$ are the degree and adjacency matrices of the pre-set knowledge graph $\mathcal{G}_{att}$, ${W}_{\mathcal{G}}^{(i)}$ denotes the parameters of $i$-th GCN layers, and $\sigma$ is an activation function. In the GCN pipeline, we assign clear attribute definitions to the initially disordered visual features $X^{(l)}$, making GCN: $\mathcal{F}_{\mathcal{G}}^{(l)}$ models them according to the relations described by the graph $\mathcal{G}_{att}$, thereby activating the implicit semantic linkage in $X^{(l)}$.

The GCN pipeline connects all GCN blocks in series, to solve the gradient diffusion on the GCN pipeline, we adopt the residual GCN structure~\cite{li2019deepgcns} shown in Figure~\ref{fig3-3}(b), as follows:
\begin{equation}
\label{eq3-7}
f_{\mathcal{G}}^{(l+1)} = F_{\mathcal{G}}^{(l+1)} \left(X^{(l+1)}, f_{\mathcal{G}}^{(l)}, <\mathcal{G}_{att}, W_{\mathcal{G}}^{(l+1)}> \right) + f_{\mathcal{G}}^{(l)}.
\end{equation}

Since we record the GCN pipeline as $\mathcal{F}_{\mathcal{G}}$, its target output is designed to the attribute word vectors of category:
\begin{equation}
\label{eq3-8}
\mathcal{A}(x_{i}) = \mathcal{F}_{\mathcal{G}} \left(x_{i}, <\mathcal{G}_{att}, W_{\mathcal{G}}> \right),
\end{equation}
where $W_{\mathcal{G}}$ is the parameters of the GCN pipeline, and
\begin{equation}
\label{eq3-9}
\mathcal{A}(x_{i}) = \left[ \mathrm{a}_{1}(x_{i}), \mathrm{a}_{2}(x_{i}), \dots, \mathrm{a}_{k}(x_{i}) \right],
\end{equation}
which is the set of attribute word vectors. $\mathrm{a}_{1}(x_{i}) \dots \mathrm{a}_{k}(x_{i})$ are the word vectors of attributes ${Att}_{1} \dots {Att}_{k}$, these attributes belong to the ground-truth category $y_{i}$.

How we get the attribute word vector: in ZSL data sets, there are several attributes belongs to each category so that we can consider each category $y \in \mathcal{Y}$ as a bag of attribute $\{ Att_{1}, Att_{2}, \dots, Att_{k} \}$, where $k$ is the number of attributes that the category $y$ has, then we can consider all categories $\mathcal{Y}$ as corpora. With this corpora, we can conveniently train a word embedding $\mathcal{M}_{e}$ with existing language model tools~\cite{mikolov2013distributed,pennington2014glove}. Like the knowledge graph $\mathcal{G}_{att}$, the acquisition of the word embedding $\mathcal{M}_{e}$ is off-line and fast. We can query the word vector $\mathrm{a}_{j} = \left[e_{j}^{1}, e_{j}^{2}, \dots, e_{j}^{d} \right] \in \mathbb{R}^{d}$ of an arbitrary attribute $Att_{j}$ from $\mathcal{M}_{e}$, where $d$ is the fixed dimension of word embedding.

The reasons why we set the target output of the GCN pipeline as above are: 1. As a model based on vocabulary co-occurrence, the word vectors can form self-consistent regression with above semantic knowledge graph $\mathcal{G}_{att}$; 2. The semantic knowledge graph $\mathcal{G}_{att}$ only provides initial attribute relational information based on the average strong attribute co-occurrence. The GCN pipeline needs to learn more personalized attribute relations with word embedding supervision.

\subsubsection{Entanglement Strategy of CNN and GCN}
\label{sec3.2.3}

Only constructed the CNN pipelines $\mathcal{F}_{conv}$ and GCN pipeline $\mathcal{F}_{\mathcal{G}}$ is not enough. We design the entanglement strategy between them based on the following motivations: 1. The modeling process of CNN and GCN pipelines needs interaction and synchronization; 2. Hierarchical visual features need to receive information on semantic relational modeling. This section introduces the dual-pipe entanglement strategy in detail from two directions: CNN to GCN and GCN to CNN.

For \textbf{CNN to GCN}, we have described that using feature maps $X^{(l)}$ from CNN block as the input of GCN block. However, the shape of $X^{(l)}$ may not be suitable to the GCN input, so we need to reshape $X^{(l)}$ first, as:
\begin{equation}
\label{eq3-10}
X^{(l)} = \mathcal{F}_{in}^{(l)} \left(X^{(l)}, W_{in}^{(l)} \right),
\end{equation}
where $\mathcal{F}_{in}^{(l)}$ is the transform function with weights $W_{in}^{(l)}$.

In addition, starting from the second GCN block $\mathcal{F}_{\mathcal{G}}^{(2)}$, the concatenation operation $\Join$ is used to merge the feature maps $X^{(l)}$ and the output $f_{\mathcal{G}}^{l-1}$ of the previous GCN block as the input. Thus we can update Equations~\ref{eq3-5} and~\ref{eq3-7} to:
\begin{equation}
\label{eq3-11}
f_{\mathcal{G}}^{(l)} = \left\{
\begin{array}{ll}
\mathcal{F}_{\mathcal{G}}^{(l)} \left( X^{(l)}, <\mathcal{G}_{att}, W_{\mathcal{G}}^{(l)}> \right) ,\quad &l = 1 \\
\mathcal{F}_{\mathcal{G}}^{(l)} \left( X^{(l)} \Join \mathcal{F}_{sq} \big( f_{\mathcal{G}}^{(l-1)} \big), \right. \quad &\multirow{2}*{$l \geq 2$} \\
\qquad \quad \left. <\mathcal{G}_{att}, W_{\mathcal{G}}^{(l)}> \right) + f_{\mathcal{G}}^{(l-1)},
\end{array}
\right.,
\end{equation}
where $\mathcal{F}_{sq}$ is a squeeze function, which reduces the dimension of $f_{\mathcal{G}}^{(l-1)}$ to $d$ for convenient, it saves the calculation of $\mathcal{F}_{\mathcal{G}}^{(l)}$.

For \textbf{GCN to CNN}, we apply the gate mechanism to feedback the semantic relationship modeling information of the GCN pipeline to the visual features of the CNN pipeline, as follows:
\begin{equation}
\label{eq3-12}
\tilde{X}^{(l)} = \sigma_{out} \left( f_{\mathcal{G}}^{(l)}, W_{out}^{(l)} \right) \otimes X^{(l)},
\end{equation}
where, the weight $W_{out}^{(l)}$ is used to align the dimension of the graph modeling information $f_{\mathcal{G}}^{(l)}$ with the feature graph $X^{(l)}$, and its role is similar to $W_{in}^{(l)}$. $\sigma_{out}$ is an activation function and outcomes value within range $[0, 1]$, operation $\otimes$ is the element-wise multiplication, and $\tilde{X}^{(l)}$ is the new visual feature maps, which will continue feed to the subsequent CNN pipeline, as: $\mathcal{F}_{conv}^{[l + 1, \sim]} \left(\tilde{X}^{(l)}, W_{\theta}^{[l + 1, \sim]} \right)$, where $[l + 1, \sim]$ meas "from $l + 1$ to the end".

The entanglement strategy of the proposed GVSE network is shown in Figure~\ref{fig3-1}, and it makes the dual-pipe structure of the GVSE network get full interaction in addition to their respective precise modeling functions. Our strategy differs from the other ZSL models based on attention methods~\cite{xie2019attentive,liu2019attribute,huynh2020shared,huynh2020fine} in particular: 1. Utilizing graph modeling to optimize visual implicit semantic information; 2. CNN features of various layers will get optimization and feedback.

The motivation for GCN semantic graph modeling on each CNN layer is that we consider the visual features of various CNN block contain implicit semantic relationship information, and modeling the semantic relationship of each CNN block supports to integrate the visual-semantic modeling into the entire CNN pipeline, which provides the semantic-rich ZSL visual embedding with complete optimization. In other words, our semantic graph modeling of visual features will not be squeezed into one block. It will be evenly distributed in the hierarchy, to share semantic modelling pressure and obtain hierarchical visual-semantic information.

\subsubsection{Integrate Hierarchical Semantic Features into Visual Embedding}
\label{sec3.2.4}

\begin{figure}[t]
\centering
\includegraphics[scale=0.6]{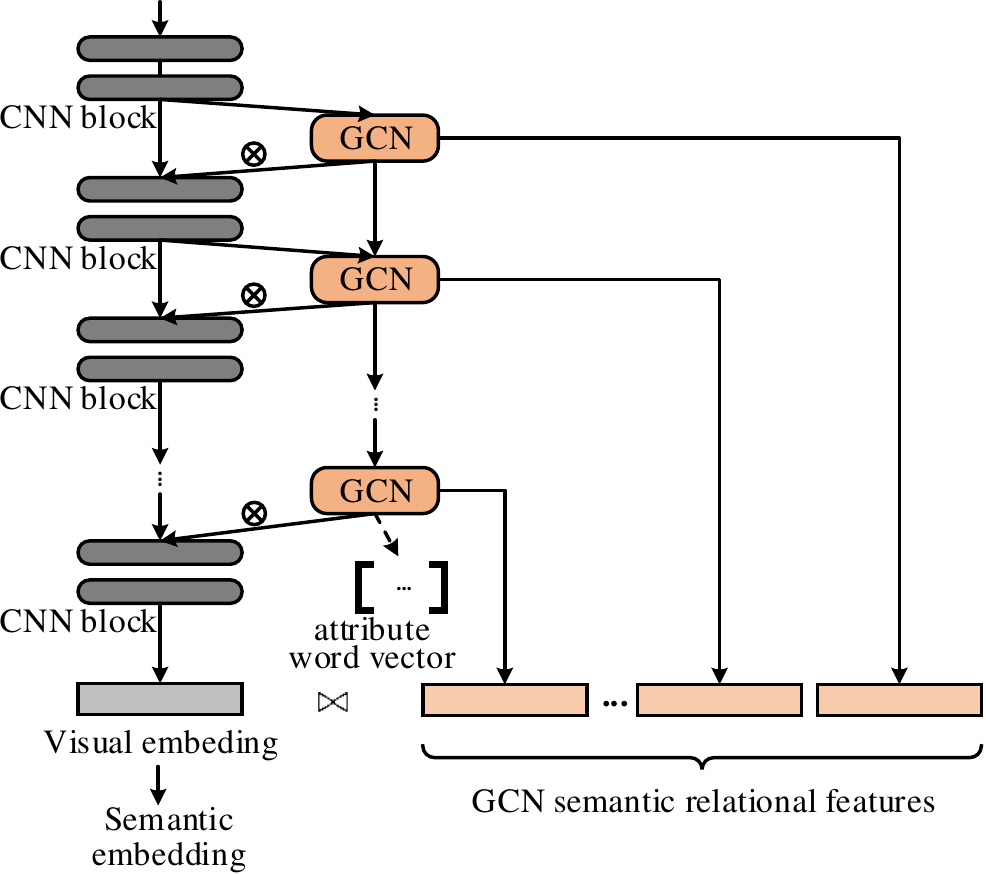}
\caption{Illustration of the GVSE network with the semantic relational features (SRFs) that fused into the final visual embedding.}
\label{fig3-4}
\end{figure}

To further enhance the final visual embedding, we perform one last step. As illustrated in Figure~\ref{fig3-3}, inspired by the pyramid structure~\cite{lin2017feature}, we merge the semantic features from GCN pipeline into the final visual embedding $\theta (x)$, i.e. \textbf{Semantic Relational Features (SRF)}. Its advantages are: 1. Help the loss back-propagation to the GCN blocks; 2. Provide more semantic information for the latent attributes~\cite{li2018discriminative,liu2019attribute}.

Note that the final visual embedding is concentrated by global average pooling as:
\begin{equation}
\label{eq3-13}
\theta(x_{i}) = \frac{1}{R \times C} \sum_{r=1}^{R} \sum_{c=1}^{C} \mathcal{F}_{conv}^{[\sim]} \left(x_{i}, W_{\theta} \right)[r, c],
\end{equation}
where $\mathcal{F}_{conv}^{[\sim]}$ is the part of CNN pipeline before global pooling, $(R \times C)$ is the shape of final feature maps of CNN pipeline, and $\mathcal{F}_{conv}^{[\sim]} \left(x_{i}, W_{\theta} \right)[r, c]$ indicates the signal of $r$-th row and $c$-th column on CNN feature maps.

The integrated visual embedding ${\theta (x)}^{+}$ is obtained as follows:
\begin{equation}
\label{eq3-14}
{\theta (x)}^{+} = \theta (x) \Join \hat{f_{\mathcal{G}}}^{(1)} \Join \hat{f_{\mathcal{G}}}^{(2)} \Join \cdots \Join \hat{f_{\mathcal{G}}}^{(L)},
\end{equation}
where
\begin{equation}
\label{eq3-15}
\hat{f_{\mathcal{G}}}^{(i)} = \frac{1}{k} \sum_{v=1}^{k} \left( \mathcal{F}_{sq} ( f_{\mathcal{G}}^{(i)}[v] ) \right),
\end{equation}
which is utilized to pool the GCN outputs of attribute word vector, and $k$ is the attribute number. Still, $\mathcal{F}_{sq}$ is a squeeze function to reduce the width of GCN outputs.

\section{Optimization Details and ZSL prediction}
\label{sec4}

The GVSE network in this study provides ZSL with a more robust feature of semantic relation representation, which is widely suitable for various ZSL frameworks. Since we have introduced the dual-pipe feedforward feature modeling method of the GVSE network in Section~\ref{sec3}, in this section, we briefly describe the ZSL prediction method with the GVSE network and the optimization method for GVSE features.

We introduce the execution of the ZSL prediction and optimization from inductive and transductive these two ZSL settings. The proposed GVSE network uses the back-propagation (BP) algorithm~\cite{rumelhart1986learning} for optimization, and the loss function indicate the detailed optimization strategy.

\begin{table*}[!htb]
\scriptsize
\begin{center}
\caption{Comparisons under the CZSL setting (\%). For each dataset, the best, the second-best, and the third-best performances are marked in \textbf{\underline{bold}}, \textbf{\textcolor{blue}{blue}}, and \textbf{\textcolor{green}{green}} font respectively for both the inductive and transductive methods. For GVSE-18, GVSE-50, and GVSE-101, the visual embedding subnet is ResNet with 18, 50, and 101 layers. Both standard split (SS) and proposed split (PS) are considered~\cite{xian2018zero}. The notations: $\mathcal{I}$ denotes inductive ZSL methods, and $\mathcal{T}$ denotes transductive ZSL methods.}
\label{table5-1}
\begin{tabular}{l|l|cc|cc|cc}
\hline
 & \multirow{2}{30pt}{\textbf{Method}} & \multicolumn{2}{c|}{\textbf{AwA2}} & \multicolumn{2}{c|}{\textbf{CUB}} & \multicolumn{2}{c}{\textbf{SUN}} \\
 \cline{3-8}
 & & SS & PS & SS & PS & SS & PS \\
\hline
\multirow{16}{10pt}[0pt]{$\mathcal{I}$}& DAP~\cite{lampert2009learning} & 58.7 & 46.1 & 37.5 & 40.0 & 38.9 & 39.9 \\
& SSE~\cite{zhang2015zero} & 67.5 & 61.0 & 43.7 & 43.9 & 25.4 & 54.5 \\
& DEVISE~\cite{frome2013devise} & 68.6 & 59.7 & 53.2 & 52.0 & 57.5 & 56.5 \\
& ESZSL~\cite{romera2015embarrassingly} & 75.6 & 55.1 & 43.7 & 53.9 & 57.3 & 54.5 \\
& ALE~\cite{akata2013label} & 80.3 & 62.5 & 53.2 & 54.9 & 59.1 & 58.1 \\
& CDL~\cite{jiang2018learning} & 79.5 & 67.9 & 54.5 & 54.5 & 61.3 & \textcolor{green}{\textbf{63.6}} \\
& AREN~\cite{xie2019attentive} & \textcolor{blue}{\textbf{86.7}} & 67.9 & \textcolor{blue}{\textbf{70.7}} & \textcolor{green}{\textbf{71.8}} & 61.7 & 60.6 \\
& LFGAA~\cite{liu2019attribute} & 84.3 & 68.1 & 67.6 & 67.6 & \textcolor{green}{\textbf{62.0}} & 61.5 \\
& TCN~\cite{jiang2019transferable} & 70.3 & 71.2 & -- & 59.5 & -- & 61.5 \\
& SGMA~\cite{zhu2019semantic} & 83.5 & 68.8 & 70.5 & 71.0 & -- & -- \\
& APNet~\cite{liu2020attribute} & -- & 68.0 & -- & 57.7 & -- & 62.3 \\
& DAZLE~\cite{huynh2020fine} & -- & -- & 67.8 & 65.9 & -- & -- \\
\cline{2-8}
& GVSE-18 (ours) & 77.3 $\pm$ 0.8 & 67.5 $\pm$ 1.1 & 68.1 $\pm$ 1.2 & 67.2 $\pm$ 0.6 & 59.2 $\pm$ 1.5 & 59.0 $\pm$ 1.1 \\
& GVSE-50 (ours) & 82.6 $\pm$ 1.6 & \textcolor{green}{\textbf{71.4 $\pm$ 1.2}} & 69.2 $\pm$ 0.9 & 70.3 $\pm$ 1.3 & 61.6 $\pm$ 1.3 & 61.7 $\pm$ 1.0 \\
& GVSE-101 (ours) & \textcolor{green}{\textbf{85.8 $\pm$ 0.8}} & \textcolor{blue}{\textbf{73.2 $\pm$ 1.0}} & \textcolor{green}{\textbf{70.6 $\pm$ 1.1}} & \textcolor{blue}{\textbf{72.2 $\pm$ 1.4}} & \textbf{\underline{64.3 $\pm$ 1.0}} & \textcolor{blue}{\textbf{63.8 $\pm$ 1.3}} \\
& GVSE$^{LT}$-101 (ours) & \textbf{\underline{88.1 $\pm$ 0.7}} & \textbf{\underline{74.0 $\pm$ 1.5}} & \textbf{\underline{71.4 $\pm$ 1.2}} & \textbf{\underline{72.8 $\pm$ 1.1}} & \textcolor{blue}{\textbf{64.2 $\pm$ 1.6}} & \textbf{\underline{64.8 $\pm$ 0.9}} \\
\hline
\hline
\multirow{12}{10pt}[-5pt]{$\mathcal{T}$} & SE-ZSL~\cite{kumar2018generalized} & 80.8 & 69.2 & 60.3 & 59.6 & 64.5 & 63.4 \\
& QFSL~\cite{song2018transductive} & 84.8 & 79.7 & 69.7 & 72.1 & 61.7 & 58.3 \\
& PREN~\cite{ye2019progressive} & 95.7 & 74.1 & 66.9 & 66.4 & 63.3 & 62.9 \\
& $f$-VAEGAN~\cite{xian2019f} & -- & \textcolor{green}{\textbf{89.3}} & -- & \textcolor{blue}{\textbf{82.6}} & -- & \textcolor{blue}{\textbf{72.6}} \\
& LFGAA + SA~\cite{liu2019attribute} & 94.4 & 84.8 & \textcolor{blue}{\textbf{79.7}} & 78.9 & 64.0 & 66.2 \\
& WDVSc~\cite{wan2019transductive} & \textcolor{green}{\textbf{96.7}} & 87.3 & 74.2 & 73.4 & 67.8 & 63.4 \\
& HPL~\cite{zhang2019hierarchical} & \textcolor{blue}{\textbf{97.8}} & \textcolor{blue}{\textbf{91.2}} & 75.3 & 75.2 & \textbf{\underline{80.4}} & \textcolor{green}{\textbf{70.4}} \\
& Zero-VAE-GAN~\cite{gao2020zero} & 93.8 & 85.4 & 69.1 & 68.9 & \textcolor{green}{\textbf{68.4}} & 66.8 \\
\cline{2-8}
& GVSE-50 + T (ours) & 91.0 $\pm$ 0.7 & 80.5 $\pm$ 0.9 & 74.2 $\pm$ 1.8 & 76.0 $\pm$ 0.9 & 60.5 $\pm$ 0.8 & 62.6 $\pm$ 1.2 \\
& GVSE-101 + T (ours) & 93.2 $\pm$ 0.9 & 83.7 $\pm$ 0.9 & \textcolor{green}{\textbf{77.6 $\pm$ 1.2}} & \textcolor{green}{\textbf{79.4 $\pm$ 1.4}} & 64.0 $\pm$ 1.5 & 63.8 $\pm$ 1.2 \\
& GVSE$^{LT}$-101 + SA (ours) & 96.3 $\pm$ 0.8 & 86.8 $\pm$ 0.9 & \textbf{\underline{81.0 $\pm$ 1.2}} & \textbf{\underline{82.9 $\pm$ 0.9}} & 65.0 $\pm$ 1.2 & 67.9 $\pm$ 1.1 \\
& GVSE-101 + WDVSc (ours) & \textbf{\underline{98.5 $\pm$ 1.1}} & \textbf{\underline{92.3 $\pm$ 1.0}} & 76.9 $\pm$ 1.7 & 76.1 $\pm$ 0.9 & \textcolor{blue}{\textbf{75.5 $\pm$ 1.3}} & \textbf{\underline{73.4 $\pm$ 0.8}} \\
\hline
\end{tabular}
\end{center}
\end{table*}

\subsection{Inductive ZSL Setting}
\label{sec4.1}

Since Eq.~\ref{eq3-4} formulates the score function of ZSL prediction, we pick the label with maximum score as:
\begin{equation}
\label{eq4-1}
y_{i}^{*} = \mathop{\arg\max}_{y \in \mathcal{Y}} p(y | x_{i}) = \mathop{\arg\max}_{y \in \mathcal{Y}} \left(\varphi(x_{i})^{\top} \phi(y) \right),
\end{equation}
where $\varphi(x_{i}) = \mathcal{F}_{\varphi} \left(\theta(x_{i})^{+}, W_{\varphi} \right)$ refers to the visual-semantic projection which is target to attributes.

We also introduce latent attributes (LA)~\cite{li2018discriminative,liu2019attribute} to supplement the imperfect artificially defined attribute space. This mechanism needs to calculate the prototypes of all classes for latent attributes, for seen classes: $\phi_{lat}(y^{s}) = \frac{1}{N} \sum_{i} \varphi_{lat}(x_{i})$, where $\varphi_{lat}(x_{i})$ denotes the discriminated features specifically for the latent attributes, which is elaborated by~\cite{li2018discriminative}. For unseen classes, the prototypes can be obtained from a ridge regression:
\begin{equation}
\label{eq4-2}
\beta_{y}^{u} = \mathop{\arg\min}_{y \in \mathcal{Y}^{\mathcal{S}}} \big\Vert \phi(y^{u}) - \sum \beta_{y}^{u} \phi(y)  \big\Vert_{2}^{2} + \big\Vert \beta_{y}^{u} \big\Vert_{2}^{2},
\end{equation}
\begin{equation}
\label{eq4-3}
\phi_{lat}(y^{u}) = \sum_{y \in \mathcal{Y}^{\mathcal{S}}} \beta_{y}^{u} \phi_{lat}(y),
\end{equation}
then, the ZSL prediction is preformed as:
\begin{equation}
\label{eq4-4}
y_{i}^{*} = \mathop{\arg\max}_{y \in \mathcal{Y}} \left(\varphi(x_{i})^{\top} \phi(y) + \varphi_{lat}(x_{i})^{\top} \phi_{lat}(y) \right),
\end{equation}

Since we have the above ZSL prediction process, the loss function is set as follows, we utilize the softmax cross-entropy loss to optimize the visual-semantic bridging, as follows:
\begin{equation}
\label{eq4-5}
\mathcal{L}_{A} = -\frac{1}{N} \sum_{i}^{N} \log \frac{\mathrm{exp} \left( \varphi (x_{i})^{\top} \phi (y_{i}) \right)}{\sum_{y \in \mathcal{Y}^{\mathcal{S}}} \mathrm{exp} \left( \varphi (x_{i})^{\top} \phi (y) \right)}.
\end{equation}

For discriminated features $\varphi_{lat}$, we follow the same strategy in~\cite{li2018discriminative,liu2019attribute} that uses triplet loss~\cite{schroff2015facenet} to enlarge the inter-class distance and reducing the intra-class distance:
\begin{equation}
\label{eq4-6}
\begin{aligned}
\mathcal{L}_{LT} = &\frac{1}{N} \sum_{i}^{N} \left[ {\Vert \varphi_{lat}(x_{i}) - \varphi_{lat}(x_{j}) \Vert}^{2} \right. \\
&  \left. \qquad - {\Vert \varphi_{lat}(x_{i}) - \varphi_{lat}(x_{r}) \Vert}^{2} + \alpha \right]_{+},
\end{aligned}
\end{equation}
where $x_{i}$, $x_{j}$, and $x_{r}$ represent the anchor, positive, and negative sample in a triplet respectively, $[\cdot]_{+}$ is equivalent to $\mathrm{max}(0, \cdot)$, $\alpha$ is the margin of triplet loss and is $1.0$ in all experiments.

As formulated by Eq~\ref{eq3-8}, since we use attribute word vectors as the target output of GCN pipeline $\mathcal{F}_{\mathcal{G}}$, the mean-square error is utilized as the loss of this branch, as:
\begin{equation}
\label{eq4-7}
\mathcal{L}_{W} = \frac{1}{N} \sum_{i}^{N} \sum_{v}^{k_{i}} { \left\Vert \mathcal{F}_{sq} \big( f_{\mathcal{G}}^{(L)}(x_{i}) \big)[v] - \mathrm{a}_{v} (y_{i}) \right\Vert }^{2},
\end{equation}
where $f_{\mathcal{G}}^{(L)}$ indicates the outcome feature from the last GCN block and $k_{i}$ is the number of attributes belongs to class $y_{i}$.

The complete loss function is as follows:
\begin{equation}
\label{eq4-8}
\mathcal{L} = \mathcal{L}_{A} + \mathcal{L}_{LT} + \gamma \mathcal{L}_{W}.
\end{equation}

\subsection{Transductive ZSL Setting}
\label{sec4.2}

We also applied the relation enhanced features learned by GVSE network to three transductive ZSL frameworks: 1. Quasi-Fully Supervised Learning (QFSL)~\cite{song2018transductive}; 2. Self-Adaptation (SA)~\cite{liu2019attribute}; 3. Wasserstein-Distance-based Visual Structure Constraint(WDVSc)~\cite{wan2019transductive}. These transduced ZSL methods all open access to unseen data to the model, thereby alleviating the "domain shift" phenomenon.

Among them, the SA method needs to be performed in conjunction with LA mechanism~\cite{li2018discriminative}; it absorbs the information of unseen category and conducts several offline iterations to update the prototype $\phi_{lat}(y^{u})$ and the attribute distribution $\phi(y^{u})$ for unseen classes.

The other two transductive ZSL methods are not utilized together with LA mechanism. In QFSL, a bias term $\mathcal{L}_{B} = \frac{1}{N^{u}} \sum_{i=1}^{N^{u}} \left( - \ln \sum_{y^{u} \in \mathcal{Y}^{\mathcal{U}}} p(y^{u} | x_{i}) \right)$ is added to the loss function to encourage the model to output the classification results of the unseen categories.

In WDVSc, the Wasserstein-distance~\cite{cuturi2013sinkhorn} is utilized to correct the position of the cluster centres of the unseen categories in the semantic transductive space. And it determines the classification result by calculating the distance between the feature distribution and the category cluster centres.

The special design of prediction and optimization in these transductive ZSL methods will not affect the modeling of the GVSE network for the relation-enhanced visual embedding ${\theta (x)}^{+}$. Their detailed implementations are described in the original studies~\cite{song2018transductive,liu2019attribute,wan2019transductive}.

\section{Experiments}
\label{sec5}

\subsection{Experimental Setup}
\label{sec5.1}

\subsubsection{Datasets}
\label{sec5.1.1}

We conduct the experiments on three representative ZSL datasets: Animals with Attribute 2~\cite{xian2018zero} (AwA2), Caltech-UCSD Birds 200-2011~\cite{WahCUB_200_2011} (CUB) and Scene Classification Database~\cite{patterson2012sun} (SUN). AwA2 is a coarse-grained dataset containing $37,322$ images from $50$ animal classes with $85$ attributes. CUB is a fine-grained dataset consisting of $11,788$ images from $200$ different bird species with $312$ attributes. SUN is another fine-grained dataset that including $14,340$ images from $717$ different scenes providing $102$ attributes. The zero-shot splits are adopted as $40/10$, $150/50$, and $645/72$ on AwA2, CUB, and SUN, respectively, for two split strategies: standard split and proposed split~\cite{xian2018zero}.

\subsubsection{Evaluation metrics}
\label{sec5.1.2}

The average per-class top-1 accuracy $\mathcal{ACC}$ is used as the primary metric. The experiments include both conventional and generalised ZSL settings~\cite{xian2018zero}. In the CZSL setting, all test samples come from the unseen classes ($y \in \mathcal{Y}^{\mathcal{U}} $). In the GZSL setting, the test samples come from both seen and unseen classes ($y \in \mathcal{Y}^{\mathcal{S}} \cup \mathcal{Y}^{\mathcal{U}} $), and we report the accuracy $\mathcal{ACC}^{\mathcal{S}}$, $\mathcal{ACC}^{\mathcal{U}}$ of seen and unseen test samples and their harmonic mean $\mathcal{H} = \frac{2 \times \mathcal{ACC}^{\mathcal{S}} \times \mathcal{ACC}^{\mathcal{U}}}{\mathcal{ACC}^{\mathcal{S}} + \mathcal{ACC}^{\mathcal{U}}}$.

\subsubsection{Implementations}
\label{sec5.1.3}

The backbone network ResNet~\cite{he2016deep} is applied to initialise the Embedding Subnet using the pre-trained weights on ImageNet-$1k$ dataset~\cite{deng2009imagenet}. Moreover, we test ResNet~\cite{he2016deep} with 18, 50, and 101 layers as a backbone to demonstrate that the proposed GVSE architecture also works well on the smaller Embedding Subnet. As far as we know, this is the first study to conduct a comparison on ZSL with different backbone scale.

For other hyper-parameters in the GVSE network, the In and Out transform functions $\mathcal{F}_{in}(\cdot)$ and $\sigma_{out}(\cdot)$ in Eqs.~\ref{eq3-10} and \ref{eq3-12} are implemented with a FC mapping, and the activation function of $\sigma_{out}(\cdot)$ is $\mathrm{Sigmoid}$. Each block in GCN pipeline contains two GCN layers with the dimension of $64$ and $R_{i} \times C_{i}$, where $(R_{i} \times C_{i})$ is the shape of the visual feature maps. The dimension $d$ of attribute word vectors is $10$. The threshold $\delta$ for edge building with measure of PMI, and balancing factor $\gamma$ are set to $0.75$ and $0.5$, respectively. We apply the Adam optimizer~\cite{kingma2014adam} throughout all experiments. Unless otherwise specified, our GVSE model utilizes the attribute word vector GCN sub-output and STFs by default.

\begin{figure}[t]
\centering
\includegraphics[scale=0.75]{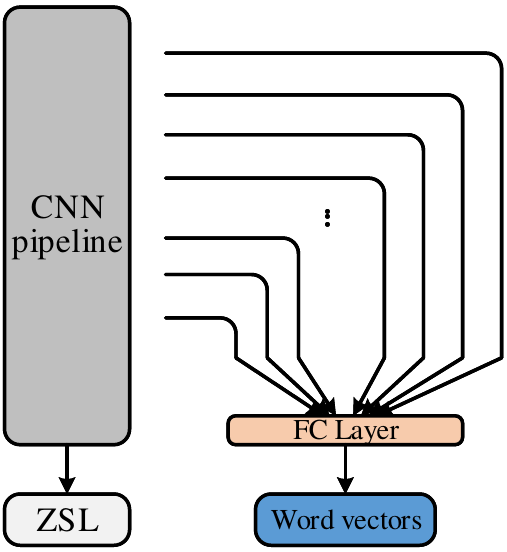}
\caption{Illustration of the "Direct" structure in ablation study on semantic modeling tools, which removes the GCN pipeline and merges the visual features of various blocks in the CNN pipeline. It utilizes the fully connected encoder to model them. The word vector auxiliary task gives the direct penalty to the outputs of each CNN block.}
\label{fig5-r1}
\end{figure}

\begin{table}[t]
\scriptsize
\begin{center}
\caption{Ablation results ($\%$) with different semantic modeling tools: 1. One FC layer direct encoder; 2. FC encoder on each block; 3. GCN pipeline (ours). ResNet-101 is utilized as backbone. Accuracy ($\mathcal{ACC}$) in CZSL with proposed split (PS) is recorded.}
\label{table5-r1}
\begin{tabular}{l|c|c|c|c}
\hline
Model & Semantic modeling tool & AwA2 & CUB & SUN \\
\hline
\multirow{3}{50pt}{ResNet-101} & Direct & 68.9 & 68.0 & 59.7 \\
& FC encoder & 69.1 & 68.4 & 61.3 \\
& GCN (ours) & \textbf{73.2} & \textbf{72.2} & \textbf{63.8} \\
\hline
\end{tabular}
\end{center}
\end{table}

\subsection{Conventional Comparison}
\label{sec5.2}

\subsubsection{Overall performance}
\label{sec5.2.1}

CZSL mainly tests the recognition ability of the model for unseen samples. We compare various state-of-the-art transductive ZSL methods~\cite{kumar2018generalized,song2018transductive,ye2019progressive,liu2019attribute,xian2019f,wan2019transductive,zhang2019hierarchical,gao2020zero} and inductive ZSL approaches~\cite{lampert2009learning,zhang2015zero,frome2013devise,romera2015embarrassingly,akata2013label,jiang2018learning,xie2019attentive,jiang2019transferable,zhu2019semantic,liu2020attribute,huynh2020fine}. In addition to an ordinary GVSE network, we also add the models of GVSE$^{LT}$ (with the latent attributes in \cite{li2018discriminative}), GVSE + T (using the transductive method in \cite{song2018transductive}), GVSE$^{LT}$ + SA (using the transductive method in \cite{liu2019attribute}) into the competition, and GVSE + WDVSc (using the transductive strategy with cluster center correcting in \cite{wan2019transductive}).

Table~\ref{table5-1} shows that the proposed GVSE network outperforms the existing ZSL methods with both inductive and transductive strategies on CZSL. In most cases, our GVSE network can achieve the best results. Even with fewer layers, the results of GVSE-50, even GVSE-18, remain extremely competitive.

In addition to the overall excellent performance, the graph-enhanced features extracted by the GVSE network can bring the considerable absolute improvements to the baselines, such as GVSE$^{LT}$-101 improves $2.2\% \sim 5.9\%$ over LFGAA, and GVSE-101 + WDVSc improves $1.8\% \sim 10.0\%$ over WDVSc. These improvements are varying, indicating that the baseline methods only perform well on individual data sets. In contrast, the GVSE model can perform well on all data sets, and the improvement is more obvious for the baseline method's poor condition. Further, The compared methods frequently only show an excellent performance on one dataset, while the GVSE can achieve the best on almost all datasets. We also found that GVSE performs well on fine-grained classification datasets such as CUB and SUN, which is most evident on the CUB datasets with detailed attribute descriptions. This indicates that the GVSE has strong cross-modal semantic modeling capabilities on fine-grained visual features and clearly described attributes.

The GVSE network does not perform as well as HPL on the standard split SUN dataset, and it has no significant advantage over $f$-VAEGAN on the CUB dataset. The reason is that when the transduction setting opens access to unseen samples, HPL and $f$-VAEGAN can provide more discriminable prototype representation space and more realistic transductive features. Fortunately, the proposed GVSE has good adaptability and does not conflict with these approaches. We plan to verify the combination of  GVSE features and more methods in the future.

\subsubsection{Comparison of the GVSE vs other attention strategies}

\begin{table}[t]
\scriptsize
\begin{center}
\caption{Ablation results with or without GVSE on proposed split (PS) (\%). For GVSE$^{\text{(last \  block)}}$, GVSE$^{\text{(each \  stage)}}$, and GVSE$^{\text{(each \  block)}}$, the GCN block (in GCN pipeline) is added for each CNN block, each stage, and the last CNN block.}
\label{table5-2}
\begin{tabular}{l|l|l|l|l}
\hline
Option & Network & AwA2 & CUB & SUN \\
\hline
\multirow{4}{50pt}{w/o GVSE} & B-Net$^{\text{w/o \ GVSE}}$ & 62.7 & 59.0 & 56.5 \\
& Att-Net~\cite{chen2017sca} & 63.9 & 64.2 & 57.7 \\
& SCA-Net~\cite{chen2017sca} & 63.3 & 65.5 & 59.0 \\
& CBAM-Net~\cite{woo2018cbam} & 62.8 & 65.8 & 58.1 \\
\hline
\multirow{3}{50pt}{w/ GVSE} & GVSE$^{\text{(last \  block)}}$ & 72.8 & 70.4 & 61.9 \\
& GVSE$^{\text{(each \  stage)}}$ & \textbf{73.2} & 71.8 & 62.9 \\
& GVSE$^{\text{(each \  block)}}$ & 72.0 & \textbf{72.2} & \textbf{63.8} \\
\hline
\end{tabular}
\end{center}
\end{table}

\begin{table*}[!htb]
\scriptsize
\begin{center}
\caption{Ablation results ($\%$) with different types of semantic graph and different types of graph modeling strategies. For semantic graph type: 1. Category similarity graph (Cate-based); 2. Instance $k$-NN graph from reference~\cite{rohrbach2013transfer} (Ins $k$-NN); 3. Attribute co-occurrence graph in the proposed approach (Att-based). For semantic modeling strategy: 1. Serial mode follow the visual modeling (Serial); 2. The proposed dual-pipeline entanglement structure (Dual-pipe). Accuracy ($\mathcal{ACC}$) in CZSL with proposed split (PS) is recorded. Results are recorded on datasets: AwA2, CUB, SUN.}
\label{table5-r2}
\begin{tabular}{l|l|c|c|c|c|c|c|c}
\hline
& \multirow{2}{50pt}{Model} & \multicolumn{3}{c|}{Type of graph} & \multirow{2}{52pt}{Modeling strategy} & \multirow{2}{15pt}{AwA2} & \multirow{2}{15pt}{CUB} & \multirow{2}{15pt}{SUN} \\
\cline{3-5}
& & Cate-based & Ins $k$-NN~\cite{rohrbach2013transfer} & Att-based & & & \\
\hline
\multirow{4}{5pt}[0pt]{$\mathcal{I}$} & \multirow{4}{50pt}{ResNet-101} & $\surd$ & -- & -- & Serial & 68.8 & 68.7 & 59.9 \\
\cline{3-9}
& & $\surd$ & -- & -- & Dual-pipe (ours) & 69.4 & 69.5 & 60.2 \\
\cline{3-9}
& & -- & -- & $\surd$ & Serial & 69.9 & 70.3 & 62.0 \\
\cline{3-9}
& & -- & -- & $\surd$ & Dual-pipe (ours) & \textbf{73.2} & \textbf{72.2} & \textbf{63.8} \\
\hline
\hline
\multirow{6}{5pt}[0pt]{$\mathcal{T}$} & \multirow{6}{50pt}{ResNet-101 + T} & -- & $\surd$ & -- & Serial & 82.2 & 77.1 & 61.0 \\
\cline{3-9}
& & -- & $\surd$ & -- & Dual-pipe (ours) & 82.0 & 76.8 & 61.9 \\
\cline{3-9}
& & $\surd$ & -- & -- & Serial & 81.4 & 76.0 & 60.8 \\
\cline{3-9}
& & $\surd$ & -- & -- & Dual-pipe (ours) & 81.8 & 76.5 & 61.4 \\
\cline{3-9}
& & -- & -- & $\surd$ & Serial & 82.4 & 77.6 & 61.3 \\
\cline{3-9}
& & -- & -- & $\surd$ & Dual-pipe (ours) & \textbf{83.7} & \textbf{79.4} & \textbf{63.8} \\
\hline
\end{tabular}
\end{center}
\end{table*}

The results of ablation study on semantic relationship modeling tools are provided. This experiment is expected to demonstrate the advantages of GCN pipelines for modeling semantic relationships within visual features. The other two options used to compare with the GCN pipeline are: 1. Remove the GCN pipeline, merge the visual features of different depths in the CNN pipeline and directly use the fully connected encoder to model them (Direct), which is illustrated as Figure~\ref{fig5-r1}; 2. The graph convolutional layers in GCN pipeline are replaced with FC layers of the same dimensions (FC encoder). The word vector auxiliary output is available here.

Tables~\ref{table5-r1} lists that the GCN pipeline has advantages for the other two FC encoders on all the data sets participating in the experiment, which confirms the positive effect of explicit graph modeling for visual-semantic linkages on category transferring. In \textbf{Appendix}, more complete results are provided.

Then, the comparison results of the proposed method and other attention models are provided. As Table~\ref{table5-2} shows, we compare the GVSE network with other attention networks to validate our graph-modeling effectiveness on ZSL. We chose the inductive version of QFSL~\cite{song2018transductive} (denoted as B-Net$^{w/o \ GVSE}$) as the baseline. To better show the benefit of GVSE, we introduce three baselines Att-Net in~\cite{chen2017sca} with spatial attention, SCA-Net~\cite{chen2017sca} and CBAM-Net~\cite{woo2018cbam} and both spatial and channel-wise attention. We also analyse the performance when adding the GCN block to each CNN block, each stage (when the size of the receptive field changes), or the last CNN block (in the \textbf{Appendix}, we present the detailed structure of different GCN block adding strategies). Although the typical attention mechanisms can improve the baseline slightly, it is still not able to compare the ability of graph modeling for a semantic representation, and the GVSE shows overwhelming advantages. Moreover, the granularity of the dataset decides requirements for GCN blocks in GVSE's GCN pipeline. On AwA2, fewer GCN blocks achieve the best results, the reason is that semantic attributes are relatively obscure in coarse-grained visual features.

\subsubsection{Ablation studies of the attribute word vector sub-output}
\label{sec5.2.3}

\begin{figure*}[t]
\centering
\includegraphics[scale=0.32]{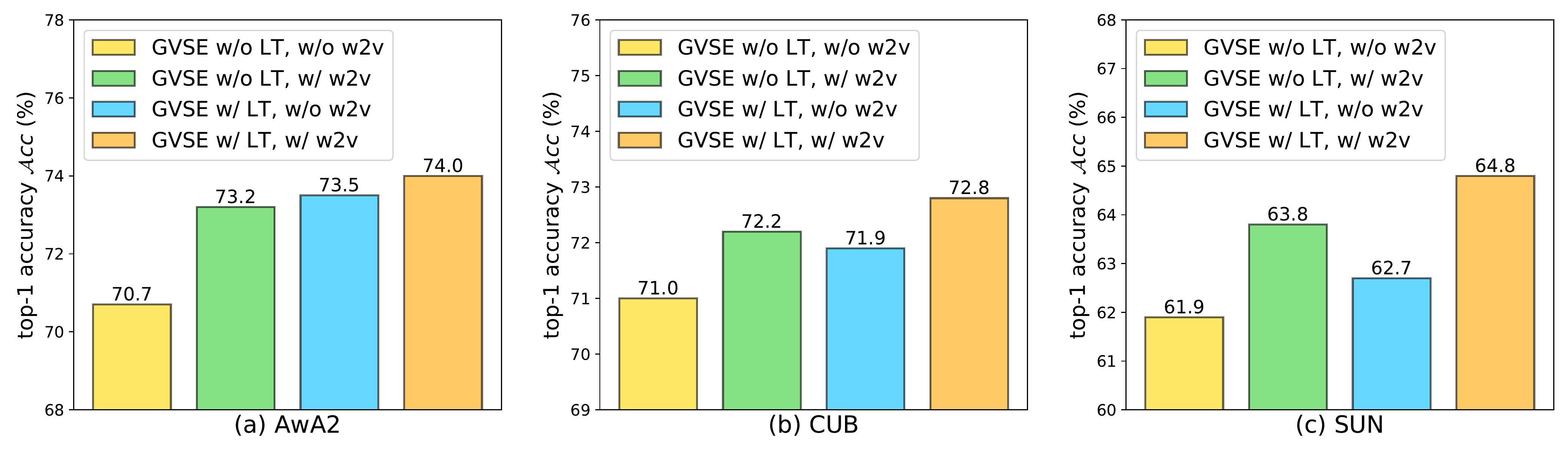}
\caption{Comparisons on AwA2, CUB, and SUN datasets with the proposed split (PS) (\%). We apply the inductive setting and set the latent attribute mechanism~\cite{li2018discriminative} and attribute word vector GCN sub-output as the ablation options.}
\label{fig5-1}
\end{figure*}

We validate the improvement from applying the attribute word vector GCN sub-output. As illustrated in Figure~\ref{fig5-1}, we can see that the regression of the semantic graph modelling by the attribute word vector target output, and this design can significantly improve the performance of visual-semantic modeling. Particularly on fine-grained image recognition datasets, like CUB and SUN, the benefits of word vector regression are higher than the latent attributes. We analyze the reason is that the feedback of fine-grained visual features on the attributes is more prominent, and the association of a fine-grained visual semantic is stronger.

Moreover, when the attribute word vector GCN sub-output and latent attributes are utilized simultaneously, the classification accuracy reaches its peak, which means that the regression of semantic graph modeling to attribute word vectors supporting ZSL model generates more accurate latent attributes.

\subsubsection{Ablation studies on using of SRF}
\label{sec5.2.4}

\begin{figure*}[t]
\centering
\includegraphics[scale=0.32]{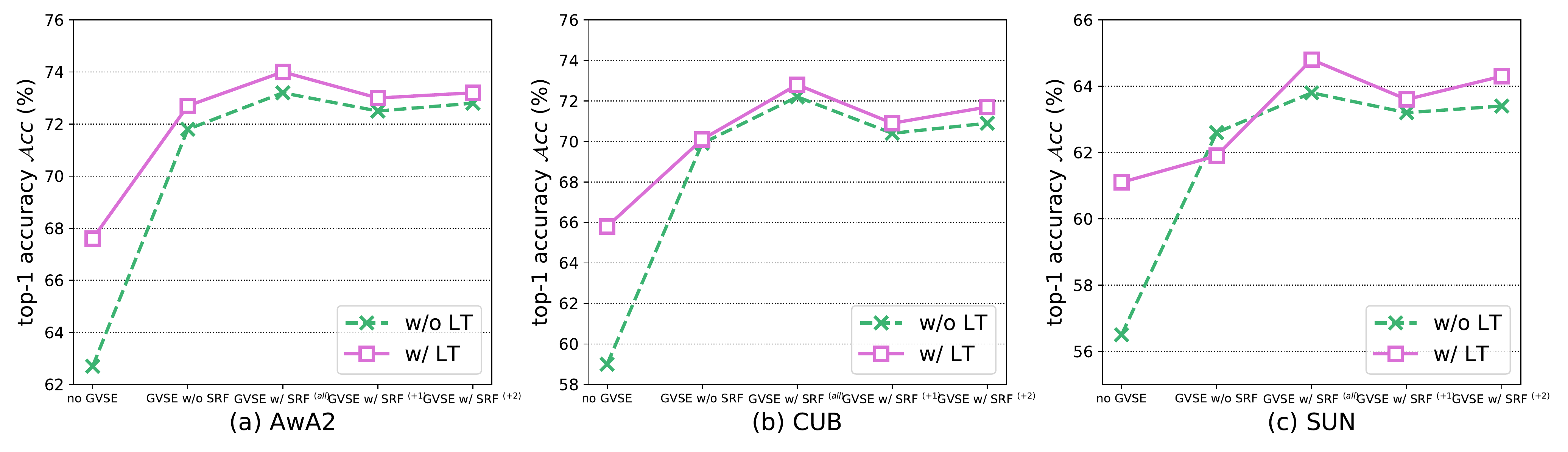}
\caption{Performance of different methods on AwA2, CUB, and SUN datasets with proposed split (PS) (\%) in the inductive setting of CZSL tasks. Here, \textbf{no GVSE} indicates baseline model with inductive ResNet QFSL~\cite{song2018transductive}, and different settings of STF are compared. Moreover, \textbf{SRF$^{(all)}$} denotes that the semantic features from all GCN blocks are merged with the visual embedding, and \textbf{SRF$^{(+1)}$} and \textbf{SRF$^{(+2)}$} indicate that only the last one or two GCN blocks are considered.}
\label{fig5-2}
\end{figure*}

We conduct the ablation study to validate the effect of SRF and set the mechanism of the latent attribute as another ablation option to test the collaborative performance of SRF and the latent attribute mechanisms~\cite{li2018discriminative,liu2019attribute}. The different structures of various SRF modes are introduced in \textbf{Appendix}. Figure~\ref{fig5-2} shows that, as SRF concatenates more semantic features into a visual embedding, the performance of the model is continuously improved, and better results are obtained when we utilize latent attributes and STF concurrently. We can conclude that SRF not only complements the visual embedding, it also boosts the modelling of the latent attributes.

In \textbf{Appendix}, we provide more comprehensive ablation research results, including all permutations and combinations of SRF feature fusion and different GCN block numbers, and some tests on residual structures' benefits for GCN pipelines.

\subsubsection{Ablation studies on the type of semantic knowledge graph}
\label{sec5.2.5}

The ablation studies on knowledge graph types and graph modeling strategies are provided in this section. For knowledge graph, there are three types of graph are introduced. In addition to the Attribute-based graph based on the attribute co-occurrence applied in the proposed method, we also apply the Category-based graph based on the cosine similarity between categories, and the instance $k$-NN graph established in reference~\cite{rohrbach2013transfer}. The cosine similarity $\mathcal{S}im$ between categories is calculated as Eq.~\ref{eq5-r1}. For the semantic graph modeling, in addition to the proposed dual-pipeline architecture, the structure of comparison is to place semantic graph modeling in series with the auxiliary output of word vector after visual modeling.

\begin{equation}
\label{eq5-r1}
\mathcal{S}im (y_{i}, y_{j}) = \frac{\phi(y_{i}) \cdot \phi(y_{j})}{\| \phi(y_{i}) \| \| \phi(y_{j}) \|},
\end{equation}
where $\phi(y_{i})$ and $\phi(y_{j})$ are the attribute distributions to categories $y_{i}$ and $y_{j}$. when $\mathcal{S}im (y_{i}, y_{j}) > 0.5$, we create an edge between categories $y_{i}$ and $y_{j}$.

As listed in Table~\ref{table5-r2}, it can be seen that when the Category-based and Instance k-NN knowledge graphs are utilized, the model's classification performance is worse than using the Attribute-based graph, demonstrating that the Attribute-based graph proposed in the GVSE model can better suit the implicit semantic relations in the visual feature. Simultaneously, the proposed Dual-pipe architecture can also achieve better results than serial modeling after visual modeling, indicating that the optimization of graph semantic modeling is demanded in all visual layers of the hierarchy. \textbf{Appendix} records the complete ablation results on the type of semantic knowledge graph.

\subsubsection{Analysis of other hyper-parameters}
\label{sec5.2.5}

\begin{table}[t]
\scriptsize
\begin{center}
\caption{Comparison results on proposed split (PS) (\%) with different setups of the PMI threshold $\delta$. The Embedding Subnet is GVSE-101.}
\label{table5-3}
\begin{tabular}{l|l|l|l|l}
\hline
Model & PMI Threshold & AwA2 & CUB & SUN \\
\hline
\multirow{3}{50pt}{GVSE-101} & $\delta = 0.25$ & 72.5 & 71.3 & 61.4 \\
& $\delta = 0.50$ & 72.2 & 71.8 & 62.1 \\
& $\delta = 0.75$ & \textbf{73.2} & \textbf{72.2} & \textbf{63.8} \\
\hline
\end{tabular}
\end{center}
\end{table}

\begin{table}
\scriptsize
\begin{center}
\caption{Comparison results on proposed split (PS) (\%) with different word embedding tools. The Embedding Subnet is GVSE-101.}
\label{tablea5-4}
\begin{tabular}{l|l|l|l|l}
\hline
Model & Word Embedding Tool & AwA2 & CUB & SUN \\
\hline
\multirow{4}{50pt}{GVSE-101} & Word2Vec~\cite{mikolov2013distributed} & 73.2 & 72.2 & \textbf{63.8} \\
& GloVe~\cite{pennington2014glove} & \textbf{73.6} & \textbf{72.4} & 63.3 \\
& FastText~\cite{bojanowski2017enriching} & 72.9 & 72.0 & 63.1 \\
\cline{2-5}
& ConceptNet~\cite{speer2017conceptnet} & 72.3 & 70.8 & 62.0 \\
\hline
\end{tabular}
\end{center}
\end{table}

\begin{table}[t]
\scriptsize
\begin{center}
\caption{Comparison results on proposed split (PS) (\%) with different SRF fusion strategies. The GVSE-101 and GVSE$^{LT}$-101 are used as the Embedding Subnet.}
\label{tablea5-5}
\begin{tabular}{l|l|l|l|l}
\hline
Model & SRF Fusion Strategy & AwA2 & CUB & SUN \\
\hline
\multirow{2}{50pt}{GVSE-101} & Summation & 72.6 & 71.9 & \textbf{64.0} \\
& Concatenation & \textbf{73.2} & \textbf{72.2} & 63.8 \\
\hline
\hline
\multirow{2}{50pt}{GVSE$^{LT}$-101} & Summation & 73.5 & 72.6 & 64.6 \\
& Concatenation & \textbf{74.0} & \textbf{72.8} & \textbf{64.8} \\
\hline
\end{tabular}
\end{center}
\end{table}

\begin{figure*}[!htb]
\centering
\includegraphics[scale=0.35]{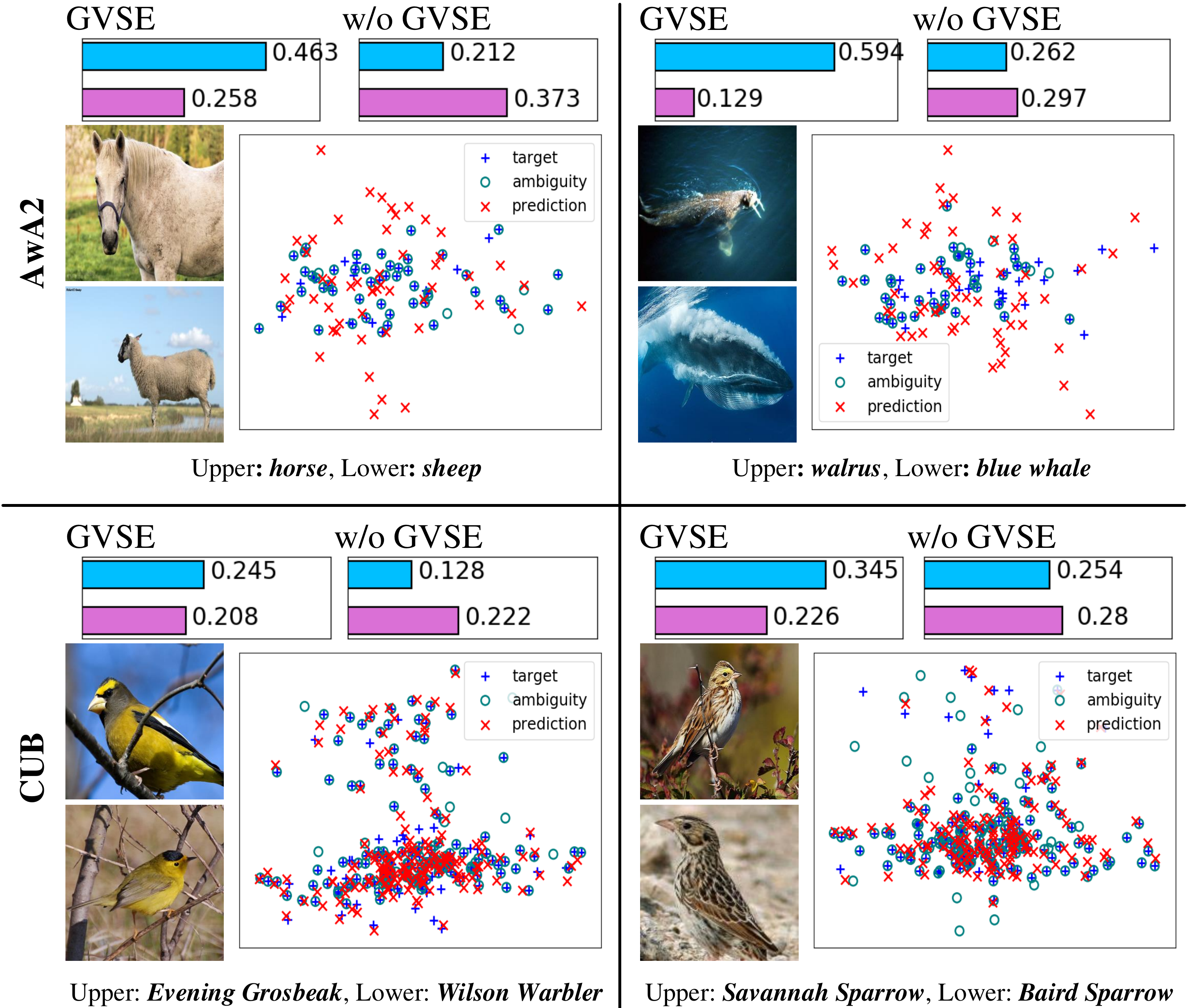}
\caption{Visualisation of the attribute word vector outcomes in CZSL task on AwA2 and CUB. The left side of each sub-figure are example images, the upper part is the input image, and the lower part is from the ambiguity class. The top of each sub-figure is the classification probabilities of the compared methods, the \textcolor{blue}{\textbf{blue}} bar refers to the target class, and \textcolor{magenta}{\textbf{pink}} indicates the ambiguity class. The symbols '\textcolor{blue}{+}', '\textcolor{green}{o}', '\textcolor{red}{x}' in the scatter represent the word vector projections of the target class, ambiguity class, and sub-output of the GVSE, respectively. The PCA dimensionality reduction~\cite{tipping1999probabilistic} is used.}
\label{fig5-3}
\end{figure*}

We first conduct experiments to check the effect the PMI threshold $\delta$, Table~\ref{fig5-3} lists the comparison ZSL results with different setting of $\delta$, it can be seen that when $\delta = 0.75$, the GVSE-101 model get the best results on all datasets. And, when $\delta = 0.25$ and $\delta = 0.5$, their gaps to the best results reached $0.7\% \sim 2.4\%$ and $0.4\% \sim 1.7\%$ respectively. Moreover, in \textbf{Appendix}, we conduct more analysis of vertex and edge Information of the semantic knowledge graph at different $\delta$ settings.

Then, we present the results of using different word embedding tools for preparing the attribute word vector GCN sub-output. As listed in Table~\ref{tablea5-4}, when we use GloVe~\cite{pennington2014glove} and FastText~\cite{bojanowski2017enriching}, and a pre-computed word embedding library based on common-sense knowledge graphs: ConceptNet Numberbatch~\cite{speer2017conceptnet}. The ZSL performance of the model changes little on GloVe and FastText. GloVe brings the best outcomes on the AwA2~\cite{xian2018zero} and CUB datasets~\cite{WahCUB_200_2011}. The model achieves the best result on the SUN dataset~\cite{patterson2012sun} with Word2Vec~\cite{mikolov2013distributed}. We do not choose a bigger language framework like BERT~\cite{devlin2019bert} because the amount of categories and attributes is tiny. In general, different self-training word embedding tools have few effects on GVSE's performance. However, when ConceptNet is utilized, the performance drops slightly. It should be noted that ConceptNet is a pre-computed word embedding package, we are only able to query attribute word, and obtain the word vector. The word vector based on common-sense knowledge graph brings more noise to ZSL image recognition in a specific field. Also, it cannot form self-consistent with the semantic graph modeling based on attribute co-occurrence in GVSE. More comprehensive results and analysis on the comparison of word embedding tools are provided in \textbf{Appendix}.

\begin{table*}[!htb]
\scriptsize
\begin{center}
\caption{Comparisons in the GZSL setting (\%). For each dataset, the best, the second best, and the third best results are marked in \textbf{bold}, \textbf{\textcolor{blue}{blue}}, and \textbf{\textcolor{green}{green}} fonts respectively. $\mathcal{I}$: inductive ZSL methods, and $\mathcal{T}$: transductive ZSL methods.}
\label{table5-6}
\begin{tabular}{l|l|ccc|ccc|ccc}
\hline
& \multirow{2}{60pt}{Method} & \multicolumn{3}{c|}{AwA2} & \multicolumn{3}{c|}{CUB} & \multicolumn{3}{c}{SUN} \\
\cline{3-11}
 & & $\mathcal{ACC}^{S}$ & $\mathcal{ACC}^{U}$ & $\mathcal{H}$ & $\mathcal{ACC}^{S}$ & $\mathcal{ACC}^{U}$ & $\mathcal{H}$ & $\mathcal{ACC}^{S}$ & $\mathcal{ACC}^{U}$ & $\mathcal{H}$ \\
\hline
\multirow{7}{5pt}[0pt]{$\mathcal{I}$} & DEVISE~\cite{frome2013devise} & \textcolor{green}{\textbf{90.5}} & 10.0 & 18.0 & \textcolor{green}{\textbf{70.9}} & 11.5 & 19.8 & \textbf{\underline{43.3}} & 7.9 & 13.4 \\
& ESZSL~\cite{romera2015embarrassingly} & 77.8 & 5.9 & 11.0 & 63.8 & 12.6 & 21.0 & 27.9 & 11.0 & 15.8 \\
& CMT*~\cite{socher2013zero} & 89.0 & 8.7 & 15.9 & 60.1 & 4.7 & 8.7 & 28.0 & 8.7 & 13.3 \\
& CDL~\cite{jiang2018learning} & 73.9 & \textcolor{green}{\textbf{29.3}} & \textcolor{green}{\textbf{41.9}} & 55.2 & 23.5 & 32.9 & 34.7 & \textcolor{green}{\textbf{21.5}} & \textcolor{green}{\textbf{26.5}} \\
& TCN~\cite{jiang2019transferable} & 65.8 & \textbf{\underline{61.2}} & \textbf{\underline{63.4}} & 52.0 & \textbf{\underline{52.6}} & \textcolor{blue}{\textbf{52.3}} & 37.3 & \textbf{\underline{31.2}} & \textbf{\underline{34.0}} \\
& LFGAA~\cite{liu2019attribute} & \textcolor{blue}{\textbf{93.4}} & 27.0 & \textcolor{green}{\textbf{41.9}} & \textcolor{blue}{\textbf{80.9}} & \textcolor{green}{\textbf{36.2}} & \textcolor{green}{\textbf{50.0}} & \textcolor{blue}{\textbf{40.0}} & 18.5 & 25.3 \\
\cline{2-11}
& GVSE-101 (ours) & \textbf{\underline{94.0 $\pm$ 1.6}} & \textcolor{blue}{\textbf{39.6 $\pm$ 2.1}} & \textcolor{blue}{\textbf{55.7 $\pm$ 1.9}} & \textbf{\underline{85.3 $\pm$ 1.3}} & \textcolor{blue}{\textbf{38.8 $\pm$ 2.2}} & \textbf{\underline{53.3 $\pm$ 1.9}} & \textcolor{green}{\textbf{39.8 $\pm$ 1.1}} & \textcolor{blue}{\textbf{24.5 $\pm$ 1.3}} & \textcolor{blue}{\textbf{30.3 $\pm$ 0.9}} \\
\hline
\hline
\multirow{12}{5pt}[0pt]{$\mathcal{T}$} & AREN~\cite{xie2019attentive} & 79.1 & 54.7 & 64.7 & 69.0 & 63.2 & \textcolor{green}{\textbf{66.0}} & \textcolor{green}{\textbf{40.3}} & 32.3 & 35.9 \\
& SGMA~\cite{zhu2019semantic} & 87.1 & 37.6 & 52.5 & 71.3 & 36.7 & 48.5 & -- & -- & -- \\
& PREN~\cite{ye2019progressive} & 88.6 & 32.4 & 47.4 & 55.8 & 35.2 & 27.2 & 35.4 & 27.2 & 30.8 \\
& LFGAA + SA~\cite{liu2019attribute} & 90.3 & 50.0 & 64.4 & \textcolor{blue}{\textbf{79.6}} & 43.4 & 56.2 & 34.9 & 20.8 & 26.1 \\
& HPL~\cite{zhang2019hierarchical} & 60.3 & \textbf{\underline{75.7}} & 67.1 & 47.1 & 54.2 & 50.4 & 39.1 & 50.9 & \textcolor{blue}{\textbf{44.2}} \\
& APNet~\cite{liu2020attribute} & 83.9 & 54.8 & 66.4 & 55.9 & 48.1 & 51.7 & \textcolor{blue}{\textbf{40.6}} & 35.4 & 37.8 \\
& E-PGN~\cite{yu2020episode} & 83.5 & 52.6 & 64.6 & 61.1 & 52.0 & 56.2 & -- & -- & -- \\
& DAZLE~\cite{huynh2020fine} & 75.7 & 60.3 & 67.1 & 59.6 & 56.7 & 58.1 & 24.3 & \textcolor{green}{\textbf{52.3}} & 33.2 \\
& Zero-VAE-GAN~\cite{gao2020zero} & 87.0 $\pm$ 3.6 & \textcolor{blue}{\textbf{70.2 $\pm$ 3.2}} & \textcolor{blue}{\textbf{77.6 $\pm$ 2.3}} & 57.9 $\pm$ 2.3 & \textcolor{green}{\textbf{64.1 $\pm$ 2.2}} & 60.8 $\pm$ 1.1 & 35.8 $\pm$ 1.0 & \textcolor{blue}{\textbf{53.1 $\pm$ 1.4}} & \textcolor{green}{\textbf{42.8 $\pm$ 0.4}} \\
\cline{2-11}
& GVSE-50 + T (ours) & \textcolor{green}{\textbf{90.8 $\pm$ 1.2}} & 65.4 $\pm$ 1.9 & \textcolor{green}{\textbf{76.0 $\pm$ 2.0}} & 69.2 $\pm$ 0.9 & \textcolor{blue}{\textbf{68.3 $\pm$ 1.1}} & \textcolor{blue}{\textbf{68.7 $\pm$ 1.3}} & 36.9 $\pm$ 1.8 & 50.6 $\pm$ 0.8 & 42.7 $\pm$ 1.7 \\
& GVSE-101 + T (ours) & \textbf{\underline{94.0 $\pm$ 1.3}} & \textcolor{green}{\textbf{69.2 $\pm$ 2.1}} & \textbf{\underline{79.7 $\pm$ 1.2}} & \textcolor{green}{\textbf{74.8 $\pm$ 1.5}} & \textbf{\underline{73.1 $\pm$ 1.3}} & \textbf{\underline{73.9 $\pm$ 1.3}} & 38.3 $\pm$ 1.2 & \textbf{\underline{55.2 $\pm$ 1.7}} & \textbf{\underline{45.2 $\pm$ 1.8}} \\
& GVSE$^{LT}$-101 + SA (ours) & \textcolor{blue}{\textbf{93.1 $\pm$ 1.3}} & 57.1 $\pm$ 1.5 & 70.8 $\pm$ 2.0 & \textbf{\underline{80.6 $\pm$ 1.6}} & 53.2 $\pm$ 1.3 & 64.1 $\pm$ 1.0 & \textbf{\underline{43.3 $\pm$ 1.5}} & 31.4 $\pm$ 1.5 & 36.4 $\pm$ 1.4 \\
\hline
\end{tabular}
\end{center}
\end{table*}

Moreover, we discuss different SRF fusion methods. In addition to using the concatenation $\Join$ to combine SRF outcomes from different GCN block, we can also use the summation function (replace Eq.~\ref{eq3-14}), as follows:
\begin{equation}
\label{eq5-1}
\theta (x)^{+} = \theta (x) \Join \sum_{i=1}^{L} \hat{f_{\mathcal{G}}}^{(i)},
\end{equation}

Using the summation to fuse all SRF features can further reduce the width of the integrated visual embedding and reduce the computational cost of followed visual-semantic bridging.

We list the comparison results with different SRF fusion strategies in Table~\ref{tablea5-5}, the concatenation method performs slightly better, and there is not much margin between the two. Among them, on the SUN dataset~\cite{patterson2012sun}, the gap is smallest, and sometimes the summation strategy is better, because the number of seen and unseen classes in the SUN dataset is hugely biased, and there are very few training samples in each category, and fewer parameters of the summation strategy can bring some benefits to generalization capabilities.

\textbf{Appendix} records more experimental results about hyper-parameter setting, include the setting of the balancing factor $\gamma$ and more results with various word embedding tools for auxiliary output.

\subsection{Generalized Comparison}
\label{sec5.3}

\begin{figure*}[!htb]
\centering
\includegraphics[scale=0.4]{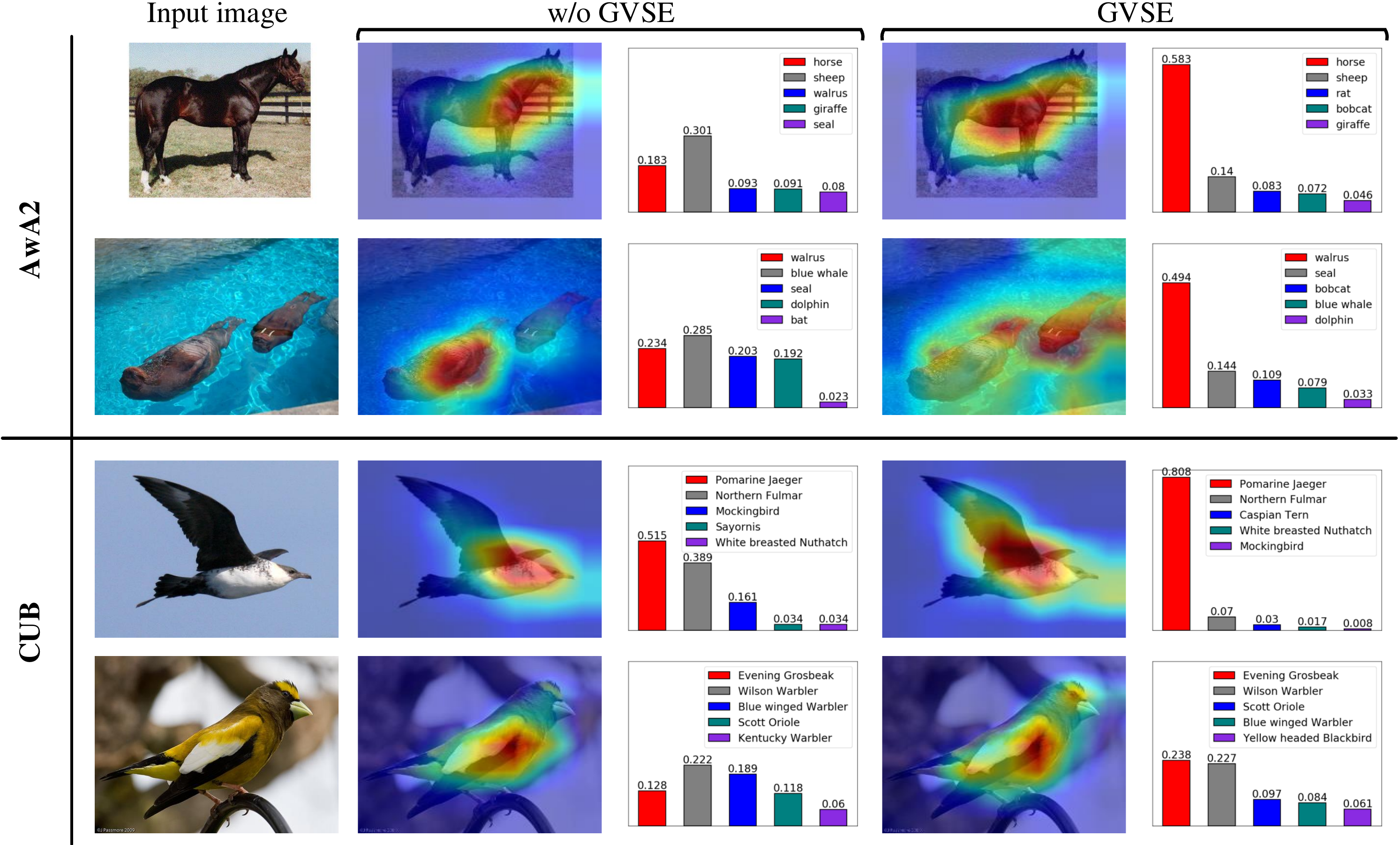}
\caption{The Grad-CAM visualization~\cite{selvaraju2017grad} and classification probabilities of GVSE$^{LT}$-101 and LFGAA (w/o GVSE). The \textcolor{red}{\textbf{red}} bar denotes the classification probability of target class, and the others indicate top-4 ambiguous classes.}
\label{fig5-4}
\end{figure*}

GZSL test model's ability to recognize samples in seen and unseen mixed search space. We also verify the proposed GVSE network in the GZSL task. As listed in Table~\ref{table5-6}, we can notice that, except for a few classification results for seen classes, our proposed method outperforms other comparison methods in the most case on all datasets. For inductive methods, GVSE-101 is lower than TCN~\cite{jiang2019transferable} in the AwA2 and SUN datasets. The reason is that although TCN does not access unseen samples when training, it utilizes the plentiful description semantic of the unseen class to help the training. For transductive methods, GVSE$^{LT}$-101 + SA surpasses state-of-the-art methods in harmonic score, and it improve over the corresponding baseline LFGAA + SA~\cite{liu2019attribute} by a significant margin of $6.4\% \sim 20.3\%$ on the AwA2, CUB, and SUN datasets. When using the transductive setting in \cite{song2018transductive}, GVSE-101 + T achieves $2.1\% \sim 7.9\%$ improvement over the comparison methods. Although the forced reversal of unseen class weights~\cite{song2018transductive} reduces the seen class accuracy on extremely biased datasets such as SUN, the GVSE-101 + T and GVSE-50 + T (with only a 50 layer backbone) still achieve the extremely competitive scores on the harmonic metric. Moreover, although many existing state-of-the-art methods utilize unseen images or semantics during training, they only achieve the best result on one or two datasets. In contrast, the GVSE network performs well on all datasets and can maintain an effective improvement when other methods fail.

Moreover, we also conduct the above ablation studies in Section~\ref{sec5.2} under the setting of GZSL, the results are provided in \textbf{Appendix}.

\subsection{Visualization and Disambiguation Analysis}
\label{sec5.4}

To gain more insightful evidence into the effectiveness of the GVSE network, we conduct visualisation experiments to show the performance of the GVSE in semantic graph modelling and visual feature attention, and analyze the disambiguation of the GVSE network.

We apply inductive GVSE$^{LT}$-101 and LFGAA to compare the models with or without the GVSE, as shown by Figure~\ref{fig5-3}. Note that the target class and the ambiguity class share numerous attributes, and thus we can see that many symbols similar to '$\oplus$' indicate the coincidence of the attributes. For each dataset, the left and right sub-figures respectively indicate strong and slightly weak ambiguities. It can be seen that a different degree of ambiguity coincide directly with the classification probabilities, and the attribute word vector output by the GVSE rarely appears near independent points of ambiguity. Moreover, the ambiguity class with a high coincidence with the target attribute space can be effectively disambiguated by the GVSE, whereas the model without an GVSE fails. For the ambiguity class with a lower coincidence, the GVSE also significantly reduces the interference of an ambiguous output.

The phenomenon shown in Figure~\ref{fig5-4} confirms that there are implicit semantic correlations between the visual features in various regions, which is in line with our expectations. The GVSE network pays attention to co-occurrence of diverse visual-semantic objects, such as (horses $\rightarrow$ muscle, meat, fields, etc), (walrus $\rightarrow$ horns, water, etc), and the head and wing of birds, while the focus area of the model w/o GVSE is limited and incomplete. From the perspective of classification probability, the proposed GVSE successfully corrects and reduces the ambiguity.

\section{Conclusion}
In this paper, we propose a Graph-based Visual-Semantic Entanglement network to help visual embedding acquire the ability to model attribute semantic relationships before performing ZSL attribute mapping. It is reasonable to give more semantic support to the visual-semantic bridging and supplement visual embedding with a semantic association representation. To accomplish this, a CNN visual modeling pipeline and a GCN relationship graph modeling modeling pipeline parallel entangled network architecture is designed. We perform an in-depth series of experiments across several datasets to validate the performance of our method, and analyse the contribution of each component utilized.

In future works, We plan further to improve the proposed model from the following aspects: 1. Explore more open semantic knowledge graph construction methods and utilize dynamic graph GNN technology for relational semantic modeling; 2. Expand the proposed method to make it applicable to a wider-ranging data condition with the word vector as the attribute space; 3. The proposed method has a wide range of applicability, and we plan to collaborate it with more ZSL models further to verify its improvement effect on more ZSL models.


%

\section*{Acknowledgment}
\footnotesize
This study was supported by China National Science Foundation (Grant Nos. 60973083 and 61273363), Science and Technology Planning Project of Guangdong Province (Grant Nos. 2014A010103009 and 2015A020217002), and Guangzhou Science and Technology Planning Project (Grant No. 201604020179, 201803010088), and Guangdong Province Key Area R \& D Plan Project (2020B1111120001).

\ifCLASSOPTIONcaptionsoff
  \newpage
\fi



\bibliographystyle{IEEEtran}
\bibliography{references}{}
%
%
%

%




\normalsize
~\\

\textbf{Yang Hu}
\textbf{(S'19)} received the MA.Eng degree from Kunming University of Science and Technology in 2016, where he is currently pursuing the Ph.D degree in South China University of Technology, China. He is also a visiting researcher in Web Science Institute at the University of Southampton, UK. His research interests include neural network and deep learning, knowledge graph, biomedical information processing.

\

\textbf{Guihua Wen}
received the Ph.D. degree in Computer Science and Engineering, South China University of Technology, and now is professor, doctoral supervisor at the School of Computer Science and Technology of South China University of Technology. His research area includes Cognitive affective computing, Machine Learning and data mining. He is also professor in chief of the data mining and machine learning laboratory at the South China University of Technology.

\

\textbf{Adriane Chapman}
is an Associate Professor in Electronics and Computer Science at the University of Southampton. She earned her PhD at the University of Michigan in 2008, and joined The MITRE Corporation, a non-profit Federally Funded Research and Development Company (FFRDC) that researched and applied solutions to technology problems in the US government. She joined the University of Southampton in 2016. She is Co-Director of the ECS Centre for Health Technologies, and is passionate about applying technology for society's benefit. She is currently working on the following projects: digital predictive technologies and conduits of algorithmic bias in policing and penal services; enabling personalized expression of privacy requirements.

\

\textbf{Pei Yang}
\textbf{(M'17)} is an associate professor at South China University of Technology. He received his Ph.D. from South China University of Technology. He has been conducting postdoctoral research in Arizona State University. His research focuses on statistical machine learning and data mining such as heterogeneous learning, semi-supervised learning, transfer learning, rare category analysis, and functional data analysis, with applications in web mining, big data analytics, bioinformatics, healthcare, etc. He has published over 30 research articles on referred journals and conference proceedings, such as TKDE, TKDD, KDD, ICDM, IJCAI, ACL, WSDM, etc. He has served on the program committee of conferences including NIPS, ICML, KDD, ICDM, SDM, IJCAI, AAAI, etc.

\

\textbf{Mingnan Luo}
Mingnan Luo is currently a master candidate in the College of Computer Science and Engineering, South China University of Technology. His main research interests include image processing and deep learning.

\

\textbf{Yingxue Xu}
is currently a master candidate in the College of Computer Science and Engineering, South China University of Technology. Her main research interests include deep learning and medical image processing.

\

\textbf{Dan Dai}
received the MA.Eng degree from Kunming University of Science and Technology in 2016, where she received her Ph.D degree in South China University of Technology, China. She is now a post-doctor research fellow in University of Lincoln, UK. Her research interests include machine learning, computer vision, and robot technology.

\

\textbf{Wendy Hall}
DBE, FRS, FREng is Regius Professor of Computer Science at the University of Southampton, and is the executive director of the Web Science Institute. She became a Dame Commander of the British Empire in the 2009 UK New Year's Honours list, and is a Fellow of the Royal Society. She has previously been President of the ACM, Senior Vice President of the Royal Academy of Engineering, and a member of the UK Prime Minister's Council for Science and Technology. She is also co-chair of the UK government's AI Review.




\end{document}